\pdfoutput=1

\documentclass[11pt]{article}

\usepackage[]{acl}

\usepackage{times}
\usepackage{latexsym}

\usepackage[T1]{fontenc}

\usepackage[utf8]{inputenc}

\usepackage{microtype}

\usepackage{amssymb}
\usepackage{graphicx}
\usepackage{tabularx}
\usepackage{amsmath}
\usepackage{latexsym}
\usepackage[normalem]{ulem}
\usepackage{subcaption}
\usepackage{xcolor}
\usepackage{multirow}
\usepackage{multicol}
\usepackage{comment}
\usepackage{inconsolata}

\title{Efficient Attribute Injection for Pretrained Language Models}
\author{
  Reinald Kim Amplayo
  \\
  University of Edinburgh \\
  \url{reinald.kim@ed.ac.uk} \And
  
  Kang Min Yoo \quad Sang-Woo Lee \\
  NAVER AI Lab, NAVER Clova AI \\
  \url{{kangmin.yoo, sang.woo.lee}@navercorp.com}
} 
\makeatletter
\newcommand{\thickhline}{%
	\noalign {\ifnum 0=`}\fi \hrule height 1pt
	\futurelet \reserved@a \@xhline
}
\makeatother

\begin{document}
\maketitle

\begin{abstract}
Metadata attributes (e.g., user and product IDs from reviews) can be incorporated as additional inputs to neural-based NLP models, by modifying the architecture of the models, in order to improve their performance.
Recent models however rely on pretrained language models (PLMs), where previously used techniques for attribute injection are either nontrivial or ineffective. 
In this paper, we propose a lightweight and memory-efficient method to inject attributes to PLMs. We extend \textit{adapters}, i.e. tiny plug-in feed-forward modules, to include attributes both independently of or jointly with the text.
To limit the increase of parameters especially when the attribute vocabulary is large, we use low-rank approximations and hypercomplex multiplications, significantly decreasing the total parameters.
We also introduce training mechanisms to handle domains in which attributes can be multi-labeled or sparse.
Extensive experiments and analyses on eight datasets from different domains show that our method outperforms previous attribute injection methods and achieves state-of-the-art performance on various datasets.
\end{abstract}

\section{Introduction}

\begin{figure}[t]
	\begin{center}
		\begin{tabular}{@{}p{7.7cm}@{}} \thickhline
			\multicolumn{1}{c}{Yelp Review}\\ \thickhline
			\textbf{Text}: My boyfriend's fav. place and the stein of beers are priced pretty good. Game nights get super packed so go early to save a seat. Kitchen closes at midnight which is too early when your buzz kicks in around 1am. \\ \hline
			\textbf{Attributes}: \\
			--- User: \texttt{n6LeAoIuDR3NfIBEsmL\_zg} \\
			--- Product: \texttt{7TMf1NuuAdvhG7IojZSKnw} \\
			\thickhline
			\multicolumn{1}{c}{Paper Abstract}\\ \thickhline
			\textbf{Text}: We present new and improved fixed-parameter algorithms for computing maximum agreement forests (MAFs) of pairs of rooted binary phylogenetic trees. The size of such a forest for two trees corresponds to their subtree prune-and-regraft distance and, if the agreement forest is acyclic, to their hybridization number ... \\ \hline
			\textbf{Attributes}: \\
			--- Authors: \texttt{Chris Whidden}, \texttt{Robert G. Beiko}, \texttt{Norbert Zeh} \\
			--- Research Areas: \texttt{q-bio.PE}, \texttt{cs.DS}
            \\\thickhline
		\end{tabular}
	\end{center}
	\caption{Examples of a Yelp review and an arXiv paper abstract and their corresponding attributes. Texts in \texttt{typewriter} font are attribute labels.}
	\label{fig:intro}
\end{figure}

Neural-based NLP models are powered by large-scale textual datasets, which are mostly crawled from the web \cite{denoyer2006wikipedia,sandhaus2008new,zhu2015aligning,ni2019justifying,raffel2020exploring}. Web texts usually are attached with \textit{metadata}, i.e. attributes
that describe the texts. For example, product reviews have user and product IDs, as well as their ratings, while research papers on arXiv have author lists and research areas as metadata attributes (see Figure \ref{fig:intro}).
While most of recent models disregard them and focus more on ungrounded language understanding (understanding language on its own, e.g., GLUE; \citealp{Wang2018GLUEAM}, \textit{inter alia}), prior work has shown that incorporating these attributes into our model increases not just its performance but also its interpretability and customizability \cite{tang2015learning,chen2016neural,kim2019categorical}.
This work explores the task of \textit{attribute injection} \cite{amplayo2019rethinking}, which aims to effectively use attributes to improve the performance of NLP models.

Previous methods \cite{tang2015learning} for attribute injection involve two steps: (a) designing an architecture that accepts both texts and attributes, and (b) training the model from scratch using task-specific datasets. \citet{chen2016neural} and subsequent work \cite{zhu2015aligning,ma2017cascading,amplayo2018cold,wu2018improving} modify the pooling module of the classifier to inject attributes, while few explore different locations such as additional memory \cite{dou2017capturing,long2018dual} and other parts of the classifier \cite{kim2019categorical,amplayo2019rethinking}.
However, these methods of modifying different modules of the model can be non-trivial when applied
to pretrained language models (PLMs; \citealp{Devlin2019BERTPO,Liu2019RoBERTaAR,qiu2020pre}). For one thing, the use of PLMs disallows designing new and specialized architectures for different domains. 
More recent work on language model customization and controllability make use of textual prompts \cite{Brown2020LanguageMA,Schick2021ItsNJ}, specialized tokens \cite{Fan2018ControllableAS,keskar2019ctrl}, and additional neural modules \cite{Wang2019HarnessingPN,liu2021gpt} to introduce additional contexts, such as style, topic, and end task.
Unfortunately, these techniques do not generalize to all kinds of attributes, such as those that are non-textual (e.g., user IDs that are not text-translatable), multi-labeled (e.g., multiple authors of a paper), and with large vocabularies (e.g., thousands of products available).


In this paper, we propose a 
method to inject attributes applicable to PLMs. Specifically, we make use of adapters \cite{Houlsby2019ParameterEfficientTL}, i.e. feed-forward modules inserted between layers of PLMs that are tiny in size, and extend them such that attributes are injected as additional inputs to the model. 
We introduce two kinds of injection methods, which either incorporate attributes independently of or jointly with the text representation.
A naive implementation of the latter would substantially increase the parameters, especially when the attribute vocabulary is large,
thus we use ideas from low-rank matrix approximations as well as parameterized hypercomplex multiplications \cite{Zhang2021BeyondFL,mahabadi2021compacter} to significantly decrease the fine-tuned parameters by up to 192$\times$ for a default base-sized BERT \cite{Devlin2019BERTPO} setting.
We also use two mechanisms, attribute dropout and post-aggregation, to handle attribute sparsity and multi-labeled attributes, respectively.
Finally, our use of adapters enables us to parameter-efficiently train our model, i.e. by freezing pretrained weights and only updating new parameters at training time.

We perform experiments on five widely used benchmark datasets for attribute injection \cite{tang2015learning,Yang2018AutomaticAP,kim2019categorical}, as well as three new datasets introduced in this paper on tasks where attributes are very important. These datasets contain attributes that have different properties (sparse vs. non-sparse, single-labeled vs. multi-labeled, etc.). Results show that our method outperforms previous approaches, as well as competitive baselines that fully fine-tune the pretrained language model. Finally, we also conduct extensive analyses to show that our method is robust to sparse and cold-start attributes and that it is modular with attribute-specific modules transferrable to other tasks using the same attributes. We
make our code and dataset publicly available.

\section{Related Work}

Prior to the neural network and deep learning era, traditional methods for NLP have relied on feature sets as input to machine learning models. These feature sets include metadata attributes such as author lists and publication venue of research papers \cite{RosenZvi2004TheAM,Joorabchi2011AnUA,Kim2017JointMO}, topics of sentences \cite{Ramage2009LabeledLA,Liu2014WebCC,Zhao2017TopicAwareDC}, as well as spatial \cite{Yang2017IdentifyingAT} and temporal \cite{Fukuhara2007UnderstandingSO} metadata attributes found in tweets. Attributes are mostly used in the area of sentiment classification \cite{Gao2013ModelingUL}, where most of the time textual data includes freely available user and product attributes. These methods rely on manually curated features that would represent the semantics of user and product information.

Deep neural networks gave rise to better representation learning \cite{Bengio2013RepresentationLA,Mikolov2013EfficientEO}, which allows us to learn from scratch semantic representation of attributes in the form of dense vectors \cite{tang2015learning}. The design of how to represent attributes has evolved from using attribute-specific word and document embeddings \cite{tang2015learning} and attention pooling weights \cite{chen2016neural,ma2017cascading,amplayo2018cold,wu2018improving}, to more complicated architectures such as memory networks \cite{dou2017capturing,long2018dual} and importance matrices \cite{amplayo2019rethinking}. These designs are model- and domain-dependent and can be non-trivial to apply to other models and datasets. Our proposed method, on the other hand, works well on any pretrained language model which are mostly based on Transformer \cite{Vaswani2017AttentionIA,Devlin2019BERTPO}.

Our work is closely related to recent literature on controlled text generation, where most of the work use either specialized control tokens concatenated with the input text \cite{Sennrich2016ControllingPI,Kikuchi2016ControllingOL,Ficler2017ControllingLS,Fan2018ControllableAS,keskar2019ctrl}, or textual prompts that instructs the model what to generate \cite{Brown2020LanguageMA,Schick2021ItsNJ,Gao2021MakingPL,Zhao2021CalibrateBU}.
While these methods have been successfully applied to pretrained language models, the attributes used to control the text are limited to those that are text-translatable (e.g., topics such as ``Technology'' or tasks that are described in text) and those with limited vocabulary (e.g., ``positive'' or ``negative'' sentiment).
In contrast, our method is robust to all kinds of attributes and performs well on all kinds of domains.

\section{Modeling Approach}

Let $x=\{x_i\}_{i=1}^N$ denote the input text of $N$ tokens, $y$ is a task-specific output, and $p(y|x)$ is a discriminative model that predicts $y$ given $x$.
Suppose there exists a set of non-textual and categorical attributes $z=\{z_j\}_{j=1}^M$ that describe text $x$ (e.g., user and product IDs of product reviews). These attributes can be multi-labeled, i.e. $z_j = [z_j^{(k)}]$ (e.g., multiple authors of a research paper) and use a finite yet possibly large vocabulary $\mathcal{Z}_j$, i.e. $z_j \subseteq \mathcal{Z}_j$.
The task of attribute injection aims to build a model $q(y|x,z)$ that additionally incorporates $z$ as input such that the difference in task performance between $p$ and $q$ is maximized. In our setting, $p$ is a pretrained language model (PLM) fine-tuned to the task, while $q$ is a PLM that also takes $z$ as additional input.

Our method can be summarized as follows. 
We extend adapters \cite{Houlsby2019ParameterEfficientTL}, which are tiny feed-forward neural networks plugged into pretrained language models, such that they also accept attributes $z$ as input.
Attributes $z$ can be represented as additional bias parameters or as perturbations to the weight matrix parameter of the adapter, motivated by how attributes are used to classify texts.
We decrease the number of parameters exponentially using low-rank matrix approximations and parameterized hypercomplex multiplications \cite{Zhang2021BeyondFL}.
Finally, we introduce two training mechanisms, attribute dropout and post-aggregation, to mitigate problems regarding attribute sparsity and multi-label properties.

The advantages of \textsc{Injectors} over previous methods are three-fold. Firstly, injecting attributes through adapters allows the model to leverage attribute information on all intermediate layers of PLMs, in contrast to previous methods where attributes can only be injected either at the beginning or at the end of PLMs.
Secondly, our use of adapters opens the possibility of parameter-efficient fine-tuning \cite{Houlsby2019ParameterEfficientTL} where only a tiny percentage of parameters is fine-tuned.
Finally, we can transfer attribute representations learned from one task to another effectively by plugging in adapters to another model.
Figure \ref{fig:model-overview} illustrates an overview of our proposed method, which we call \textsc{Injectors}.

\subsection{Preliminary: Adapters}

We first briefly describe adapters. Let $\mathbf{h} \in \mathbb{R}^{d_h}$ be the output hidden vector from a multi-head attention or feed-forward layer in a Transformer block. An adapter layer is basically two feed-forward networks that projects $\mathbf{h}$ into vector $\mathbf{h}'\in \mathbb{R}^{d_h}$  with a much smaller dimension $d_a \ll d_h$:
\begin{align}
    \label{eq:adapter}
    \mathbf{h}' &= \text{Adapt}(\mathbf{h}) \nonumber \\
    &= \text{FFNet}_{up}(f(\text{FFNet}_{down}(\mathbf{h}))) + \mathbf{h}
\end{align}
where $\text{FFNet}(\mathbf{x}) = \mathbf{Wx} + \mathbf{b}$, $\mathbf{W}$ and $\mathbf{b}$ are learned weight and bias parameters of $\text{FFNet}$, $f(\cdot)$ is a non-linear function, and the addition represents a residual layer.

Adapters are inserted every after multi-head attention and feed-forward layers for all Transformer blocks. PLMs with adapters are trained such that only the adapter parameters are updated while the original pretrained weights are left untouched. This makes training more efficient memory-wise compared to fully fine-tuning \cite{Houlsby2019ParameterEfficientTL} and more robust towards different hyperparameter settings \cite{Han2021RobustTL}.

\subsection{Our Method: Injectors}

\begin{figure}[t]
    \centering
    \includegraphics[width=\columnwidth]{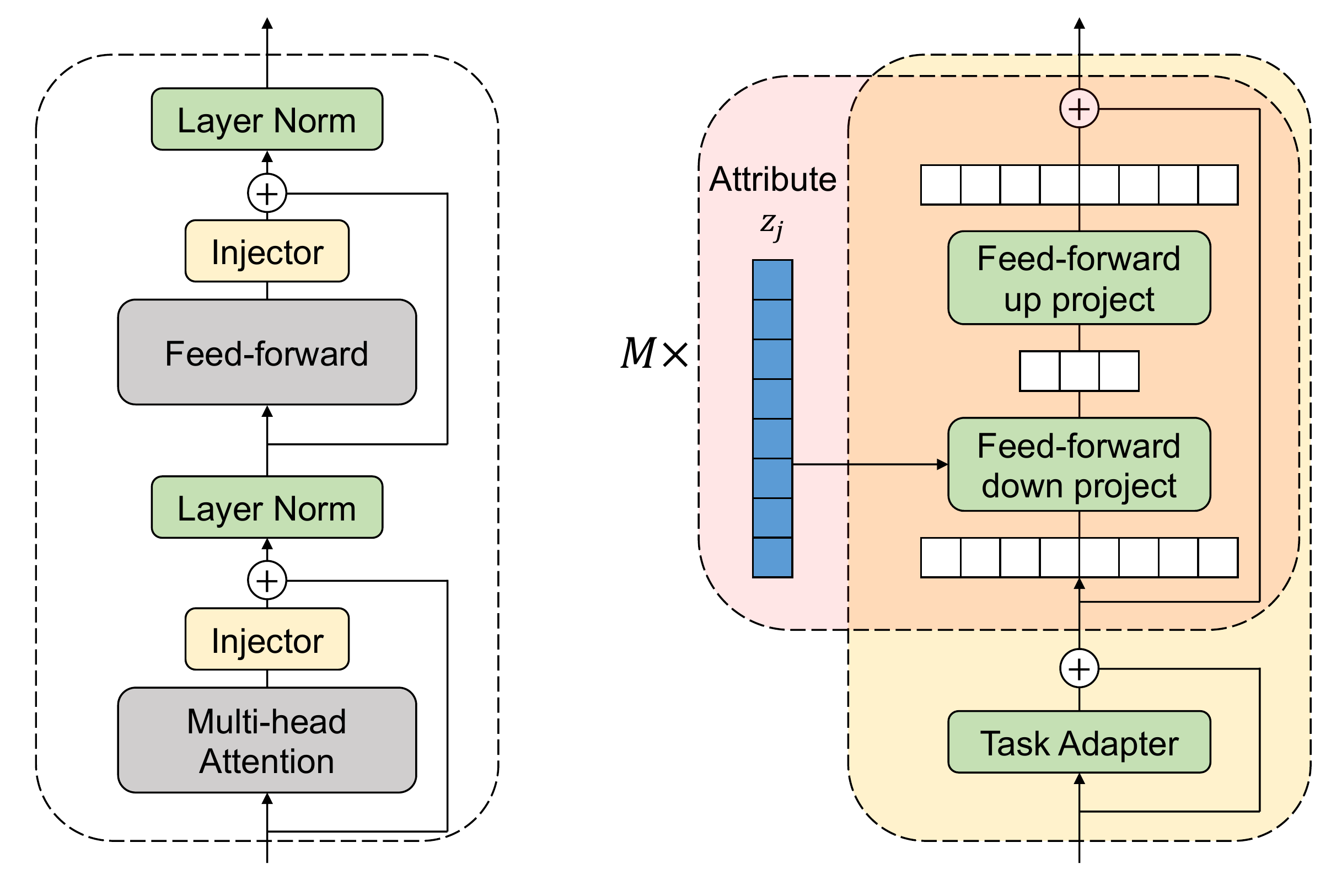}
    \caption{Architecture of the \textsc{Injector} module when integrated into one block of a Transformer model (see left of figure). \textsc{Injector} starts with a task-specific adapter, followed by $M$ attribute-specific adapters, one for each attribute given in the task (see right of figure). Green-colored modules are trained and fine-tuned, while gray-colored modules are fixed.}
    \label{fig:model-overview}
\end{figure}

In this section, we describe our method \textsc{Injectors} in detail. \textsc{Injectors} are multi-adapter modules that transforms hidden vector $\mathbf{h}$ into attribute-injected hidden vector $\mathbf{h}_z$. These are inserted right after the multi-head attention and feed-forward layers of the pretrained language model, as shown in Figure \ref{fig:model-overview}.

\paragraph{Task-specific Adapter}

\textsc{Injectors} start with a task-specific adapter that uses Equation \ref{eq:adapter} to transform the previous hidden vector $\mathbf{h}$ to $\mathbf{h}'$. The use of a separate task-specific adapter is essential to make our method modularizable and learned attributes on one task transferrable to another. We show extensive analyses on the modularity of our method in the later sections.

\paragraph{Attribute-specific Adapters}

Attributes $z$ are injected through attribute-specific adapters, where they are used in two different ways.
Firstly, they are used as \textit{bias} parameters independent of the text representation. This is motivated by the fact that attributes can have prior disposition regardless of what is written in the text. For example, a user may tend to give lower review ratings than average.
Secondly, they are also used as \textit{weight} parameters. This allows our method to jointly model attributes with the text representation. 
This is motivated by how attributes can change the semantics of the text. For example, one user may like very sweet food while another user may dislike it, thus the use of the word \textit{sweet} in the text may mean differently for them.

More formally, for each attribute $z_j \in z$, we sequentially transform the previously attribute-injected vector $\mathbf{h}_{z_{j-1}}$ to attribute-injected vector $\mathbf{h}_{z_j}$ using the following equation:
\begin{align}
    \label{eq:att-adapter}
    \mathbf{h}_{z_j} &= \text{AttrAdapt}(\mathbf{h}_{z_{j-1}}, z_j) \\ &= \text{FFNet}_{up}(f(\mathbf{W}_{z_j} \mathbf{h}_{z_{j-1}} + \mathbf{b}_{z_j})) \nonumber \\ 
    &\quad + \mathbf{h}_{z_{j-1}}
\end{align}
where $\mathbf{h}_{z_0} = \mathbf{h}'$ from the output of task-specific adapter. Unlike standard adapters, the attribute-specific weight matrix $\mathbf{W}_{z_j} \in \mathbb{R}^{d_h \times d_a}$ and bias parameter $\mathbf{b}_{z_j} \in \mathbb{R}^{d_a}$ of the down-project feed-forward network
are not learned from scratch, but instead are calculated as follows.

The calculation of the bias parameter $\mathbf{b}_{z_j}$ is trivial; we perform a linear transformation of the attribute embedding $\mathbf{z}_j$:
\begin{equation}
    \label{eq:att-bias}
    \mathbf{b}_{z_j} = g_{bias} ( \mathbf{z}_j ) + \mathbf{c}_{bias}
\end{equation}
where $g_{bias} \in \mathbb{R}^{d_z} \mapsto \mathbb{R}^{d_a}$ is a linear projection, $\mathbf{c}_{bias} \in \mathbb{R}^{d_a}$ is a learned vector, and $d_z$ is the attribute embedding size.

We also define $\mathbf{W}_{z_j}$ as:
\begin{equation}
    \mathbf{W}_{z_j} = g_{weight} ( \mathbf{z}_j ) + \mathbf{C}_{weight}
\end{equation}
where $\mathbf{C}_{weight}\in \mathbb{R}^{d_h \times d_a}$ is a learned matrix. The function $g_{weight}$, however, cannot be defined similarly as a linear projection. This would require a tensor parameter of size $d_z \times d_h \times d_a$ to linearly project $\mathbf{z}_j$ to $\mathbf{W}_j$. Considering the fact that we may have multiple attributes for each domain, the number of parameters would not scale well and makes the model very large and difficult to train. 
Inspired by \citet{mahabadi2021compacter}, we use ideas from low-rank matrix decomposition and parameterized hypercomplex multiplications (PHMs; \citealp{Zhang2021BeyondFL}) to substantially decrease the number of parameters.

\begin{figure*}[t]
    \centering
    \includegraphics[width=\textwidth]{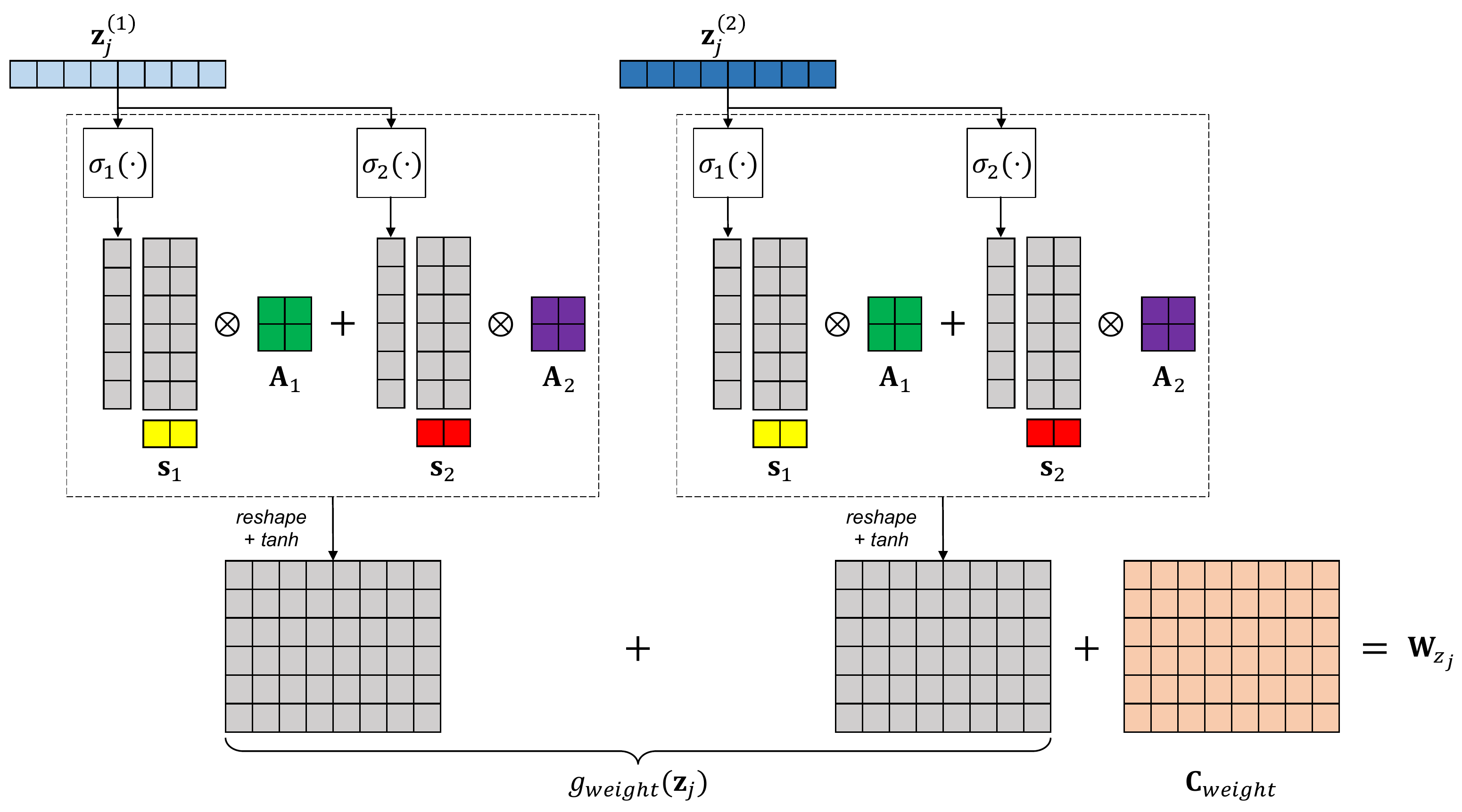}
    \caption{An illustration of how attribute embedding $\mathbf{z}_j$ is transformed into weight matrix $\mathbf{W}_{z_j}$.
    The colored tensors are learned parameters, while the gray ones are derived. By using a set of tiny parameters $\mathbf{A}_o$ and $\mathbf{s}_o$, we are able to obtain large matrices. When there are multiple labels for attribute $z_j$, we process them separately and aggregate the resulting large matrices.}
    \label{fig:lowrank}
\end{figure*}

Specifically, we first transform attribute embedding $\mathbf{z}_j$ into vectors in hypercomplex space with $O$ dimensions, i.e.:
\begin{equation}
    \label{eq:hypercomplex}
    \hat{\mathbf{z}}_j = [\sigma_1(\mathbf{z}_j),...,\sigma_O(\mathbf{z}_j)] \in \mathbb{H}^{d_z}
\end{equation}
where $\sigma_o(\cdot) \in \mathbb{R}^{d_z} \mapsto \mathbb{R}^{d_a}$ is a linear projection in the $o$th dimension. A hypercomplex vector with $O$ dimensions is basically a set of vectors with one real vector and $O-1$ ``imaginary''
vectors.\footnote{Following \citet{Tay2019LightweightAE} and \citet{Zhang2021BeyondFL}, we remove the imaginary units of these vectors to easily perform operations on them, thus these vectors are also in the real space.}

For each dimension $o$, we first define a small rank-one matrix $\mathbf{S}_o \in \mathbb{R}^{d_a \times d_h/O^2}$ as an outer product between $\hat{\mathbf{z}}_{j,o}$ and a learned vector $\mathbf{s}_o \in \mathbb{R}^{d_h/O^2}$:
\begin{equation}
    \label{eq:low-rank}
    \mathbf{S}_o = \hat{\mathbf{z}}_{j,o} \mathbf{s}_o^\top
\end{equation}
and then define a large matrix $\mathbf{\hat{W}}_{j,o} \in \mathbb{R}^{d_h \times d_a}$ as the Kronecker product, denoted by $\otimes$ between two matrices $\mathbf{S}_o$ and a learned matrix $\mathbf{A}_o \in \mathbb{R}^{O \times O}$, followed by a reshape and the hyperbolic tangent function:
\begin{equation}
    \label{eq:phm1}
    \mathbf{\hat{W}}_{o} =\text{Reshape}(\tanh( \mathbf{S}_o \otimes \mathbf{A}_o))
\end{equation}

Finally, we add the large matrices $\mathbf{\hat{W}}_{o}$ of each dimension. To sum up, 
we define $g_{weight}(\mathbf{z}_j)$ as:
\begin{align}
    \label{eq:phm2}
    &g_{weight}(\mathbf{z}_j) = \nonumber \\
    &\quad \sum\nolimits_{o=0}^O \text{Reshape}(\tanh( \sigma_o ( \mathbf{z}_j ) \mathbf{s}_o^\top \otimes \mathbf{A}_o))
\end{align}
and Figure \ref{fig:lowrank} shows an illustration.

Low-rank (Eq. \ref{eq:low-rank}) and PHMs (Eqs. \ref{eq:phm1}-\ref{eq:phm2}) are both necessary to achieve a high performance with decreased parameters. 
While low-rank in itself reduces the most parameters, it also reduces the expressive power of the model since it outputs rank-one matrices.
PHMs mitigate this by performing a sum of Kronecker products, increasing the rank of the matrix to potentially at most $O^2$.
Finally, this process effectively reduces the number of parameters from $\mathcal{O}(d_z * d_h * d_a)$ to $\mathcal{O}(d_z*d_a)$, 
since the parameters in $\sigma_o$ dominate the other parameters (see Appendix for a detailed parameter analysis).
%

\paragraph{Attribute Dropout and Post-Aggregation}

For cases where attributes are sparse and multi-labeled, we use the following mechanisms. Firstly, we add a dropout mechanism that randomly masks out attributes from training instances with a rate $r_{drop}$. This replicates how instances at test time would look like, where some attributes are not found in the vocabulary.

Secondly, when there are more than one labels of an attribute, instead of aggregating them first before processing, as in \citet{kim2019categorical}, we perform aggregation post hoc (as also shown in Figure \ref{fig:lowrank} for $\mathbf{W}_{z_j}$), i.e.:
\begin{align}
    \label{eq:post-agg}
    \mathbf{b}_{z_j} &= \sum\nolimits_k g_{bias} (\mathbf{z}_j^{(k)}) + \mathbf{c}_{bias} \\
    \mathbf{W}_{z_j} &= \sum\nolimits_k g_{weight} (\mathbf{z}_j^{(k)}) + \mathbf{C}_{weight}
\end{align}
Aggregating attribute embeddings reduces their individual representation power, while our post-aggregation mechanism preserves this since a sum of non-linear transformations is injective \cite{Xu2019HowPA}.

\section{Experimental Setup}

\paragraph{Datasets}

\begin{table*}[t]
    \small
    \centering
    \begin{tabular}{@{}lrrrrrcrrc@{}}
        \thickhline
        Dataset & \#Train & \#Dev & \#Test & \#Words/Input & \#Classes & \#Attrs & \#Attr. Vocab & \%Sparse & Multi-label? \\
        \thickhline
        Yelp 2013 & 62.5K & 7.8K & 8.7K & 210 $\pm$ 166 & 5 & 2 & 3.3K & 0.0\%& \\
        Yelp 2014 & 183.0K & 22.7K & 25.4K & 218 $\pm$ 175 & 5 & 2 & 9.0K & 0.0\% &\\
        IMDB & 67.4K & 8.4K & 9.1K & 425 $\pm$ 278 & 10 & 2 & 2.9K & 0.0\% & \\
        AAPR & 33.5K & 2.0K & 2.0K &  97 $\pm$ 36  & 2 & 2 & 51.6K & 97.8\% & \checkmark \\
        PolMed & 4.5K & \multicolumn{1}{c}{--} & 0.5K & 38 $\pm$ 62 & 9 & 4 & 0.5K & 63.8\% \\
        \hline
        Food.com & 162.4K & 20.3K & 20.3K & 101 $\pm$ 65 & 16 & 3 & 40.5K & 80.0\% & \checkmark \\
        Goodreads & 714.7K & 10.0K & 10.0K & 132 $\pm$ 72 & 2 & 3 & 43.8K & 34.4\% \\
        Beeradvocate & 1.5M & 10.0K & 10.0K & 133 $\pm$ 56 & 4 $\times$ 9 & 3 & 98.0K & 75.3\% \\
        \thickhline
    \end{tabular}
    \caption{Dataset statistics. The second block reports new datasets introduced in this paper. Beeradvocate is a multi-task dataset, with nine classes for each of four given aspects. \%Sparse is the percentage of attributes with less than 10 training examples. Multi-label attributes include lists of authors and research areas for AAPR, and lists of ingredients and tags for Food.com.}
    \label{tab:data-stats}
\end{table*}

We performed experiments on a total of eight datasets. Five of them are widely used datasets for attribute injection, which include the following: 
\begin{enumerate}
    \item \textbf{Yelp 2013} \cite{tang2015learning}: A review rating prediction dataset where we are tasked to predict the rating of a review given two attributes, the user and the product.
    \item \textbf{Yelp 2014} \cite{tang2015learning}: A dataset similar to Yelp 2013, but larger in size.
    \item \textbf{IMDB} \cite{tang2015learning}: A dataset similar to Yelp 2013, but with ten rating scales and longer reviews.
    \item \textbf{AAPR} \cite{Yang2018AutomaticAP}: A dataset for classifying whether an arXiv paper is accepted to a conference or not, with two attributes, a list of authors and a list of research areas.
    \item \textbf{PolMed} \cite{kim2019categorical}: A message type classification dataset in the political domain, where the goal is to classify a tweet into one of nine classes, with four attributes, the politician who wrote the message, the media source, the audience, and the political bias.
\end{enumerate}

We also introduce three new benchmark datasets with larger size, where attributes  are crucial to effectively solve the task, are sparser, and have large vocabularies.
\begin{enumerate}
    \setcounter{enumi}{5}
    \item \textbf{Food.com} \cite{Majumder2019GeneratingPR}: A dataset where given a recipe of a food and three attributes, the user, a list of ingredients, and a list of tags, we are tasked to predict the estimated number of minutes it takes to make the food, rounded down to the tens.
    \item \textbf{Goodreads} \cite{Wan2018ItemRO}: A spoiler prediction dataset where we classify whether a book review contains spoiler or not, with three attributes, the user, the book, and the rating of the review.
    \item \textbf{Beeradvocate} \cite{McAuley2012LearningAA}: A multi-aspect rating prediction dataset where given a beer review and three attributes, the user, the beer, and the overall review rating, we are tasked to predict the ratings of four \textit{aspects}, or properties that influence user satisfaction, of the beer: appearance, aroma, palate, and taste.
\end{enumerate}

Table \ref{tab:data-stats} reports statistics of all the datasets.

\paragraph{Training Configuration}

For our PLM, we used weights and settings of \texttt{bert-base-uncased} \cite{Devlin2019BERTPO}, available in the HuggingFace library \cite{Wolf2020TransformersSN}.
We set the dimensions of all parameters as follows: $d_z = d_h = 768$, $d_a = 64$, and $O = 4$.
Using this setting and our parameter-saving method, we are able to decrease the parameters by $192\times$ the naive method.
We set both the general and attribute dropout rates to $0.2$ and the batch size to $8$. We used Adam with weight decay \cite{Loshchilov2019DecoupledWD} to optimize our models with a learning rate of $3e-5$ and $200K$ training steps, with the first $20K$ steps used to warm-up training linearly.

To train our models, we added a logistic classifier which
transforms the \texttt{[CLS]} token into logits. The weights here are updated during training.
We then used a cross entropy loss to train the models on all datasets except for Goodreads and Beeradvocate.
The Goodreads dataset is very imbalanced towards the negative class (i.e., not spoiler). We thus put more importance to detecting the spoiler class and used a weighted cross entropy loss with $0.5$ weight on the negative class and $1.0$ weight on the positive class.
For Beeradvocate, we treat the task as a multi-task problem, where each aspect rating prediction is a separate task. Thus, we used multiple classifiers, one for each aspect, and aggregate the losses from all classifiers by averaging.
For PolMed where there is no available development set, we performed 10-fold cross-validation, following \citet{kim2019categorical}.

\paragraph{Comparison Systems}

We compared our method with several approaches, including the following no-attribute baselines:
\begin{enumerate}
    \item {BERT-base} ($\mathcal{B}$): The base model used in our experiments.
    \item $\mathcal{B}$ + \textsc{Adapters}: Extra tiny parameters are added to the base model and are used for training instead of the full model.
\end{enumerate}

Baselines with attributes injected include the following models. We use the same base model $\mathcal{B}$ for all baselines for ease of comparison.

\begin{enumerate}
    \setcounter{enumi}{2}
    \item $\mathcal{B}$ + \textsc{Tokens}: Following work from controlled text generation \cite{Sennrich2016ControllingPI}, the attributes are used as special control tokens prepended in front of the input.
    \item $\mathcal{B}$ + \textsc{UPA}: Short for User-Product Attention \cite{chen2016neural}, attributes are used as additional bias vectors when calculating the weights of the attention pooling module.
    \item $\mathcal{B}$ + \textsc{CHIM}: Short for Chunk-wise Importance Matrices \cite{amplayo2019rethinking}, attributes are used as importance matrices multiplied to the weight matrix of the logistic classifier.
\end{enumerate}

Finally, we also included in our comparisons the state-of-the-art from previous literature whenever available.

\section{Experiments}

\paragraph{Main Results}

\begin{table*}[t]
    \small
	\centering
	\begin{tabular}{@{}lcccccccc@{}}
		\thickhline
		\multicolumn{1}{c}{Model} &  Yelp 2013    & Yelp 2014    & IMDB & AAPR & PolMed & Food.com & Goodreads & BeerAdvocate \\
		\thickhline
		BERT-base ($\mathcal{B}$) & 67.97 & 68.07 & 48.10 & 63.70 & 41.82 & 41.89 & 41.98 & 50.48 \\
		$\mathcal{B}$ + \textsc{Adapters} & 66.47 & 67.44 & 46.41 & 62.85 & 44.24 & 42.02 & 48.92  & 50.71\\
		\hline
		$\mathcal{B}$ + \textsc{Tokens} & \textcolor{red}{67.87} & \textcolor{red}{67.98} & \textcolor{red}{48.00} & 64.85 & 42.63 & \textcolor{red}{41.23} & 44.79 & \textcolor{red}{50.25} \\
		$\mathcal{B}$ + \textsc{UPA} & 68.38 & 68.82 & 48.90 & 64.40 & 42.83 & 43.97 & 43.96 & 51.98 \\
		$\mathcal{B}$ + \textsc{CHIM} & 68.71 & 68.56 & 49.36 & 65.30 & 43.64 & 43.35 & 43.58 & 52.29 \\
		$\mathcal{B}$ + \textsc{Injectors} (ours) & \hspace*{.15cm}\textbf{70.76}${}^*$ & \hspace*{.15cm}\textbf{71.35}${}^*$ & \hspace*{.15cm}55.13${}^*$ & \textbf{67.10} & \hspace*{.15cm}\textbf{47.27}${}^*$ & \textbf{45.01} & \hspace*{.15cm}\textbf{57.78}${}^*$ & \hspace*{.15cm}\textbf{57.69}${}^*$ \\ \hline
		Prev. SOTA & \hspace*{-.17cm}67.8 & \hspace*{-.17cm}69.2 & \hspace*{-.17cm}\textbf{56.4} & 66.15 & 41.89 & -- & -- & -- \\
		\thickhline
	\end{tabular}%
	\caption{Performance (F1-score on Goodreads, Accuracy otherwise)  of competing methods on the eight datasets. The first block includes PLMs without injected attributes, while the second block includes those with injected attributes. 
	Attribute injected PLMs that perform worse than the base model $\mathcal{B}$ are colored red.
	Numbers for previous SOTA are copied from \citet{amplayo2019rethinking} for Yelp and IMDB datasets and from \citet{kim2019categorical} otherwise.
	Best systems are shown in bold. Asterisk (*) means there is a significant difference between our model and the CHIM model (paired bootstrap resampling; $p < 0.05$).}
	\label{tab:results}%
\end{table*}%

We evaluated system outputs with accuracy for all datasets except Goodreads, where we used F1-score. For brevity in Beeradvocate, we took the average of the accuracy of all sub-tasks. Our results are summarized in Table \ref{tab:results}. The non-attribute baselines perform similarly except in PolMed and Goodreads, where \textsc{Adapters} improve the base model significantly, which aligns to previous findings on the robustness of adapters \cite{Han2021RobustTL}.
Overall, \textsc{Injectors} outperforms all baselines on all datasets. When compared with the previous state-of-the-art, our model outperforms on all datasets except IMDB. We account the performance decrease on the length limit of PLMs since most reviews in IMDB are longer than 512 words (see Table \ref{tab:data-stats}, not considering subwords), whereas the previous state-of-the-art \cite{amplayo2019rethinking} used a BiLSTM with no length truncation as base model.

\paragraph{Ablation Studies}

\begin{table}[t]
    \small
	\centering
	\begin{tabular}{@{}lccc@{}}
		\thickhline
		\multicolumn{1}{c}{Model} & \textsc{Food} & \textsc{Good} & \textsc{Beer} \\
		\thickhline
		$\mathcal{B}$ + \textsc{Injectors} & {45.01} & {57.78} & {57.69} \\ \hline
		\quad $-$ bias injection & 44.86 & 57.06 & 57.67 \\
		\quad $-$ weight injection & 44.74 & 57.61 & 57.29 \\ 
		\quad $-$ task adapter & 44.30 & 56.57 & 57.28 \\ \hline
		\quad $-$ attribute drop & 44.48 & 56.62 & 57.41 \\
		\quad $-$ post-aggregation & 43.30 & 57.78 & 57.69 \\ \hline
		\quad $-$ low-rank & OOM & OOM & OOM \\
		\quad $-$ PHM & 43.89 & 55.99 & 56.51 \\
		\thickhline
	\end{tabular}%
	\caption{Performance on \textsc{Food}.com, \textsc{Good}reads, and \textsc{Beer}Advocate of \textsc{Injectors} and versions thereof without some of our proposed components (second block), training mechanisms (third block), and parameter-saving methods (fourth block). OOM denotes the model does not run on our experimental setup due to out of memory error.}
	\label{tab:ablation}%
\end{table}%

We present in Table \ref{tab:ablation} various ablation studies on the three new datasets (see Appendix for the other datasets), which assess the contribution of different model components. Our experiments confirm that the use of both bias and weight injection as well as the addition of task adapter improve performance.
Interestingly, some datasets prefer one injection type over the other. Goodreads, for example, prefers bias injection, that is, using attributes as prior and independent of the text (e.g., the tendency of the user to write spoilers).
Moreover, our training mechanisms also increase the performance of the model. This is especially true for post-aggregation on the Food.com dataset since two of its attributes are multi-labeled (ingredients and tags).
Finally, we show that, on their own, the parameter-saving methods either perform worse or do not run at all.

\paragraph{On Attribute Sparsity}

\begin{figure*}
    \centering
    \includegraphics[width=0.32\textwidth]{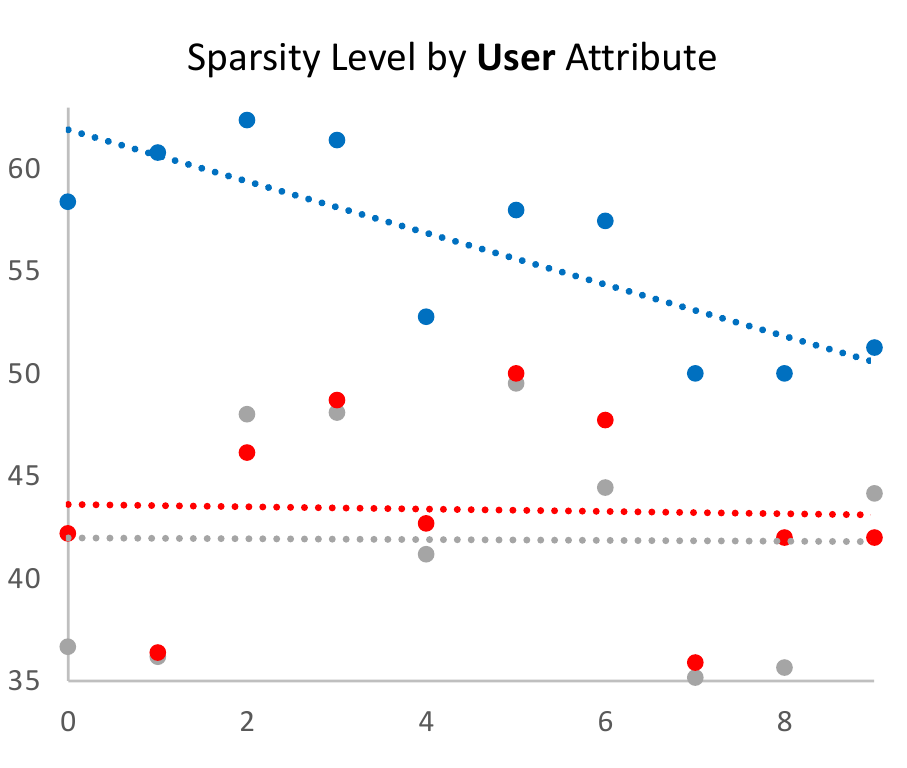}
    \includegraphics[width=0.32\textwidth]{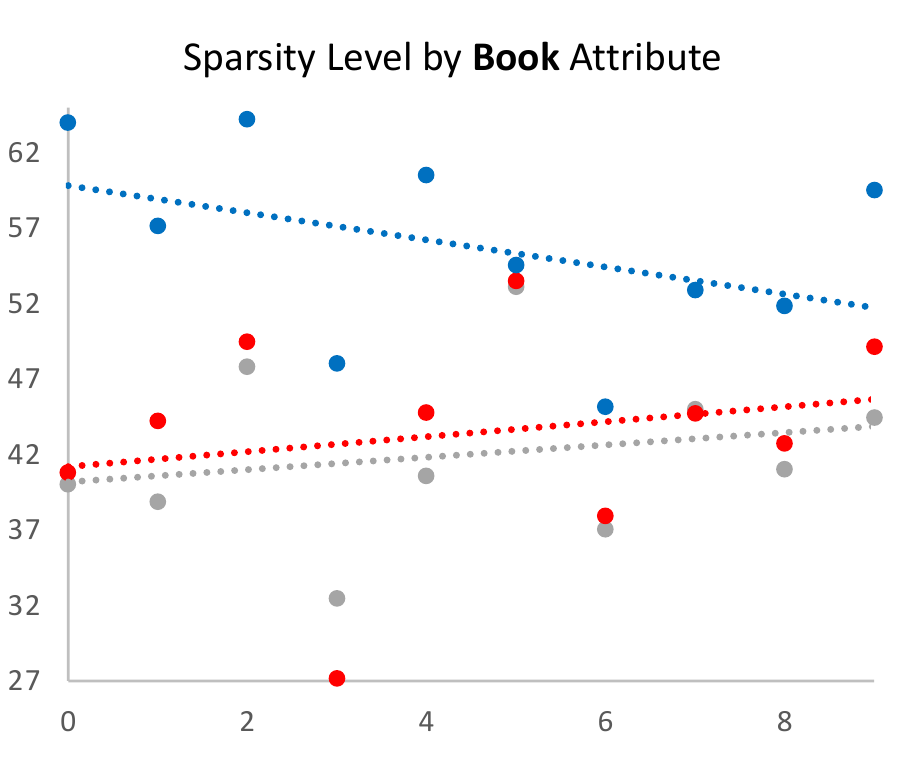}
    \includegraphics[width=0.32\textwidth]{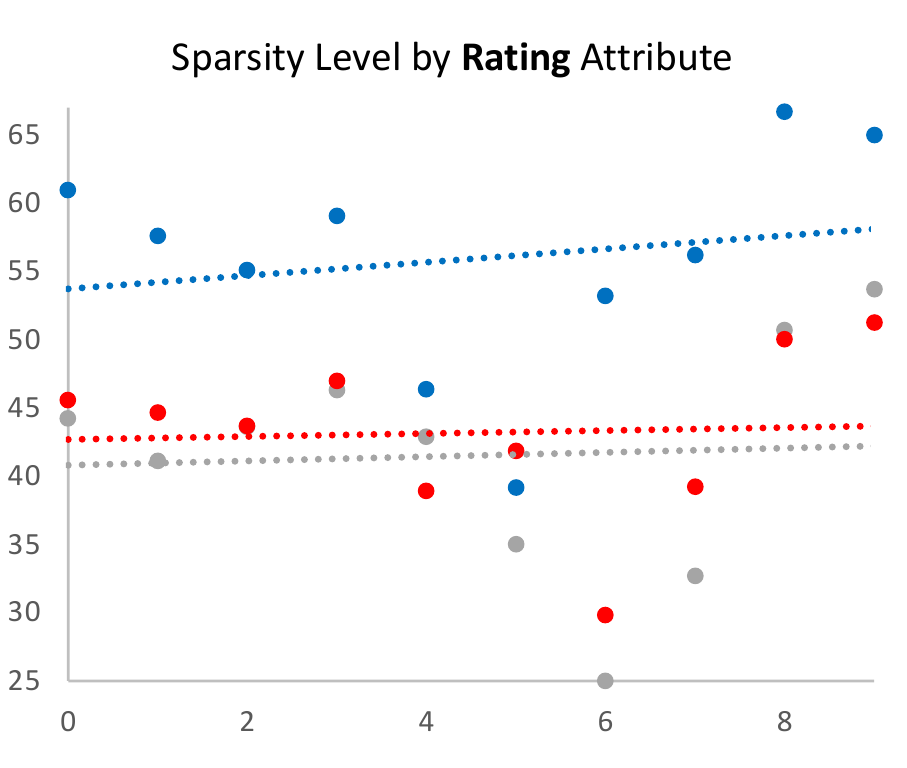}
    \caption{Performance plots per sparsity level of BERT-base (gray), CHIM (red), and \textsc{Injectors} (blue) for each attribute in the Goodreads dataset. The x-axis with a lower value has a higher sparsity level (0 is the most sparse).}
    \label{fig:sparse-graph}
\end{figure*}


\begin{table}[t]
    \small
	\centering
	\begin{tabular}{@{}lccc@{}}
		\thickhline
		& \multicolumn{3}{c@{}}{Yelp 2013} \\
		\multicolumn{1}{c}{Model} & 20\% sparse & 50\% sparse & 80\% sparse \\
		\thickhline
		BERT-base ($\mathcal{B}$) & 64.69 & 62.65 & 56.83 \\
		$\mathcal{B}$ + \textsc{CHIM} & \textcolor{red}{64.63} & \textcolor{red}{62.43} & \textcolor{red}{56.01} \\
		$\mathcal{B}$ + \textsc{Injectors} & \textbf{67.22} & \textbf{63.51} & \textbf{58.14} \\ \hline
		HCSC & \hspace*{-.17cm}63.6 & \hspace*{-.17cm}60.8 & \hspace*{-.17cm}53.8 \\
		\thickhline
		\\
		\thickhline
		& \multicolumn{3}{c@{}}{IMDB}  \\
		\multicolumn{1}{c}{Model} & 20\% sparse & 50\% sparse & 80\% sparse \\
		\thickhline
		BERT-base ($\mathcal{B}$) & 43.49 & 37.45 & 29.82  \\
		$\mathcal{B}$ + \textsc{CHIM} & \textcolor{red}{43.24} & \textcolor{red}{35.72} & 31.48 \\
		$\mathcal{B}$ + \textsc{Injectors} & \textbf{50.74} & 43.68 & 34.02 \\\hline
		HCSC & \hspace*{-.17cm}50.5 & \hspace*{-.17cm}\textbf{45.6} & \hspace*{-.17cm}\textbf{36.8} \\
		\thickhline
	\end{tabular}%
	\caption{Performance on the sparse versions of Yelp 2013 and IMDB. Best systems are shown in bold. Models perform worse than $\mathcal{B}$ are colored red.}
	\label{tab:sparse}%
\end{table}%

As shown in Table \ref{tab:data-stats}, most datasets contain attributes that are sparse. In this section, we analyze the ability of \textsc{Injectors} on sparse attributes using two experiments. In our first experiment, we looked at the model performance at different {sparsity levels}. That is, for each attribute in the dataset, we equally divided the test set into ten bins arranged according to the attribute sparsity.
Figure \ref{fig:sparse-graph} shows attribute-specific plots of the performance for each bin of the base model, CHIM, and \textsc{Injectors} on the Goodreads dataset (see Appendix for the plots on all datasets). For sparse attributes such as user and book, the performance difference of \textsc{Injectors} from the base model increases as the sparsity level increases, showing that our method mitigates attribute sparsity well.
CHIM, on the other hand, has a uniform performance increase all throughout.

In our second experiment, we checked the performance of the models when trained using synthetically created sparse versions of Yelp 2013 and IMDB provided in \citet{amplayo2018cold}, where the datasets are downsampled such that the attributes become 20/50/80\% more sparse than the original.
We compared the performance of BERT-base, CHIM, and \textsc{Injectors}.
Table \ref{tab:sparse} shows their performance on the datasets, along with HCSC \cite{amplayo2018cold}, which is a BiLSTM- and CNN-based model with a sparsity-aware attribute injection method similar to the UPA method \cite{chen2016neural}.
As can be seen, our method still performs the best on these datasets, while CHIM underperforms and is worse than the base model on all cases except on IMDB 80\% sparse.
Our method performs better than HCSC on Yelp 2013, and competitively on IMDB where input texts are longer than 512 tokens.

\paragraph{On Model Modularity}


\begin{table}[t]
    \small
    \centering
    \begin{tabular}{@{}lcccc@{}}
        \thickhline
        \multicolumn{1}{c}{Model} & A & RPT$\rightarrow$A & R & APT$\rightarrow$R \\
        \thickhline
        \textsc{CHIM} & 52.6 & 52.5 (-0.2\%) & 51.3 & 51.2 (-0.2\%) \\
        \textsc{Injectors} & 56.3 & 56.0 (-0.5\%) &  54.0 & 54.5 (+0.9\%) \\
        \thickhline
    \end{tabular}
    \caption{Performance of models on single-task Beeradvocate (A: appearance, R: aroma, P: palette, T: taste). 
    The arrow ($\rightarrow$) indicates that the attribute-specific adapters of the model in the right-hand side (i.e., A and R) are initialized using parameters of the left-hand side model (i.e., RPT and APT) and are frozen during training.}
    \label{tab:modularity}
\end{table}

Since \textsc{Injectors} are basically a sequence of adapters, which are known to be self-contained modular units \cite{Pfeiffer2020AdapterHubAF}, modular composition across different models is also effective in our setting.
We verify this using the following experiment. We first divide the Beeradvocate dataset, which is a multi-aspect rating prediction task with four aspects, into two subsets:
(1) a single-task \textit{target} dataset and (2) a 3-task \textit{source} dataset.
We train our model using the source dataset, obtaining attribute-specific parameters.
We then transfer these fixed parameters when training the model using the target dataset, and only fine-tune parameters of the task adapter and the classifier.

We arbitrarily chose the first two aspects alphabetically, appearance (A) and aroma (R), as target tasks. We split the training dataset into four parts, one for each aspect, to remove biases from overlapping training datasets. We combined the non-target datasets as the source dataset. We experimented with CHIM and \textsc{Injectors}, and report the results in Table \ref{tab:modularity}.
When compared to the same model trained directly on the target task (see A and R columns), both methods are able to achieve very minimal performance loss (see RPT$\rightarrow$A and APT$\rightarrow$R columns),
with a marginal increase on the R target task when using \textsc{Injectors}. 
This confirms the results of previous work on the transferability of attribute embeddings in CHIM \cite{amplayo2019rethinking}, as well as the modularity of adapter-based modules \cite{Pfeiffer2020AdapterHubAF}.

\section{Conclusions}

We considered the use of attributes as additional contexts when fine-tuning PLMs for NLP tasks. We proposed the \textsc{Injector} module, an extension of adapters that also accepts attributes as input. Our method considers two kinds of injection strategies, uses parameter-saving techniques, and introduces training mechanisms to account for sparse and multi-labeled attributes.
Experiments on eight datasets of various classification tasks showed that our method improves substantially over previous methods.
Finally, we conducted extensive analyses on how \textsc{Injectors} handle attribute sparsity and to verify their modularity.
In the future, we plan to apply our methods to real world data where there are millions of attributes. We also plan to explore the use of attribute injection methods to text generation tasks, i.e. injecting attributes when generating texts instead of during modeling.

\section*{Acknowledgments}

We would like to thank Jaewook Kang and other members of the NAVER AI Lab for their insightful comments. Reinald is supported by a Google PhD Fellowship.

\bibliography{acl}

\begin{thebibliography}{54}
\expandafter\ifx\csname natexlab\endcsname\relax\def\natexlab#1{#1}\fi

\bibitem[{Amplayo(2019)}]{amplayo2019rethinking}
Reinald~Kim Amplayo. 2019.
\newblock Rethinking attribute representation and injection for sentiment
  classification.
\newblock In \emph{EMNLP-IJCNLP}, pages 5602--5613.

\bibitem[{Amplayo et~al.(2018)Amplayo, Kim, Sung, and Hwang}]{amplayo2018cold}
Reinald~Kim Amplayo, Jihyeok Kim, Sua Sung, and Seung-won Hwang. 2018.
\newblock Cold-start aware user and product attention for sentiment
  classification.
\newblock In \emph{ACL}, pages 2535--2544.

\bibitem[{Bengio et~al.(2013)Bengio, Courville, and
  Vincent}]{Bengio2013RepresentationLA}
Yoshua Bengio, Aaron~C. Courville, and Pascal Vincent. 2013.
\newblock Representation learning: A review and new perspectives.
\newblock \emph{IEEE Transactions on Pattern Analysis and Machine
  Intelligence}, 35:1798--1828.

\bibitem[{Brown et~al.(2020)Brown, Mann, Ryder, Subbiah, Kaplan, Dhariwal,
  Neelakantan, Shyam, Sastry, Askell, Agarwal, Herbert-Voss, Krueger, Henighan,
  Child, Ramesh, Ziegler, Wu, Winter, Hesse, Chen, Sigler, Litwin, Gray, Chess,
  Clark, Berner, McCandlish, Radford, Sutskever, and
  Amodei}]{Brown2020LanguageMA}
Tom~B. Brown, Benjamin Mann, Nick Ryder, Melanie Subbiah, J.~Kaplan, Prafulla
  Dhariwal, Arvind Neelakantan, Pranav Shyam, Girish Sastry, Amanda Askell,
  Sandhini Agarwal, Ariel Herbert-Voss, Gretchen Krueger, T.~Henighan,
  R.~Child, A.~Ramesh, Daniel~M. Ziegler, Jeff Wu, Clemens Winter, Christopher
  Hesse, Mark Chen, Eric Sigler, Mateusz Litwin, Scott Gray, Benjamin Chess,
  Jack Clark, Christopher Berner, Sam McCandlish, Alec Radford, Ilya Sutskever,
  and Dario Amodei. 2020.
\newblock Language models are few-shot learners.
\newblock In \emph{NeurIPS}.

\bibitem[{Chen et~al.(2016)Chen, Sun, Tu, Lin, and Liu}]{chen2016neural}
Huimin Chen, Maosong Sun, Cunchao Tu, Yankai Lin, and Zhiyuan Liu. 2016.
\newblock Neural sentiment classification with user and product attention.
\newblock In \emph{EMNLP}, pages 1650--1659.

\bibitem[{Denoyer and Gallinari(2006)}]{denoyer2006wikipedia}
Ludovic Denoyer and P.~Gallinari. 2006.
\newblock The wikipedia xml corpus.
\newblock In \emph{INEX}.

\bibitem[{Devlin et~al.(2019)Devlin, Chang, Lee, and
  Toutanova}]{Devlin2019BERTPO}
J.~Devlin, Ming-Wei Chang, Kenton Lee, and Kristina Toutanova. 2019.
\newblock Bert: Pre-training of deep bidirectional transformers for language
  understanding.
\newblock In \emph{NAACL}.

\bibitem[{Dou(2017)}]{dou2017capturing}
Zi-Yi Dou. 2017.
\newblock Capturing user and product information for document level sentiment
  analysis with deep memory network.
\newblock In \emph{EMNLP}, pages 521--526.

\bibitem[{Fan et~al.(2018)Fan, Grangier, and Auli}]{Fan2018ControllableAS}
Angela Fan, David Grangier, and Michael Auli. 2018.
\newblock Controllable abstractive summarization.
\newblock In \emph{NMT@ACL}.

\bibitem[{Ficler and Goldberg(2017)}]{Ficler2017ControllingLS}
Jessica Ficler and Y.~Goldberg. 2017.
\newblock Controlling linguistic style aspects in neural language generation.
\newblock \emph{ArXiv}, abs/1707.02633.

\bibitem[{Fukuhara et~al.(2007)Fukuhara, Nakagawa, and
  Nishida}]{Fukuhara2007UnderstandingSO}
T.~Fukuhara, H.~Nakagawa, and T.~Nishida. 2007.
\newblock Understanding sentiment of people from news articles: Temporal
  sentiment analysis of social events.
\newblock In \emph{ICWSM}.

\bibitem[{Gao et~al.(2021)Gao, Fisch, and Chen}]{Gao2021MakingPL}
Tianyu Gao, Adam Fisch, and Danqi Chen. 2021.
\newblock Making pre-trained language models better few-shot learners.
\newblock In \emph{ACL-IJCNLP}.

\bibitem[{Gao et~al.(2013)Gao, Yoshinaga, Kaji, and
  Kitsuregawa}]{Gao2013ModelingUL}
W.~Gao, Naoki Yoshinaga, Nobuhiro Kaji, and M.~Kitsuregawa. 2013.
\newblock Modeling user leniency and product popularity for sentiment
  classification.
\newblock In \emph{IJCNLP}.

\bibitem[{Han et~al.(2021)Han, Pang, and Wu}]{Han2021RobustTL}
Wenjuan Han, Bo~Pang, and Ying~Nian Wu. 2021.
\newblock Robust transfer learning with pretrained language models through
  adapters.
\newblock In \emph{ACL}.

\bibitem[{Houlsby et~al.(2019)Houlsby, Giurgiu, Jastrzebski, Morrone,
  de~Laroussilhe, Gesmundo, Attariyan, and
  Gelly}]{Houlsby2019ParameterEfficientTL}
N.~Houlsby, A.~Giurgiu, Stanislaw Jastrzebski, Bruna Morrone, Quentin
  de~Laroussilhe, Andrea Gesmundo, Mona Attariyan, and S.~Gelly. 2019.
\newblock Parameter-efficient transfer learning for nlp.
\newblock In \emph{ICML}.

\bibitem[{Joorabchi and Mahdi(2011)}]{Joorabchi2011AnUA}
Arash Joorabchi and A.~Mahdi. 2011.
\newblock An unsupervised approach to automatic classification of scientific
  literature utilizing bibliographic metadata.
\newblock \emph{Journal of Information Science}, 37:499 -- 514.

\bibitem[{Keskar et~al.(2019)Keskar, McCann, Varshney, Xiong, and
  Socher}]{keskar2019ctrl}
Nitish~Shirish Keskar, Bryan McCann, Lav~R Varshney, Caiming Xiong, and Richard
  Socher. 2019.
\newblock Ctrl: A conditional transformer language model for controllable
  generation.
\newblock \emph{arXiv preprint arXiv:1909.05858}.

\bibitem[{Kikuchi et~al.(2016)Kikuchi, Neubig, Sasano, Takamura, and
  Okumura}]{Kikuchi2016ControllingOL}
Yuta Kikuchi, Graham Neubig, Ryohei Sasano, Hiroya Takamura, and M.~Okumura.
  2016.
\newblock Controlling output length in neural encoder-decoders.
\newblock In \emph{EMNLP}.

\bibitem[{Kim et~al.(2019)Kim, Amplayo, Lee, Sung, Seo, and
  Hwang}]{kim2019categorical}
Jihyeok Kim, Reinald~Kim Amplayo, Kyungjae Lee, Sua Sung, Minji Seo, and
  Seung-won Hwang. 2019.
\newblock Categorical metadata representation for customized text
  classification.
\newblock \emph{TACL}, 7:201--215.

\bibitem[{Kim et~al.(2017)Kim, Kim, and Oh}]{Kim2017JointMO}
Jooyeon Kim, Dongwoo Kim, and Alice~H. Oh. 2017.
\newblock Joint modeling of topics, citations, and topical authority in
  academic corpora.
\newblock \emph{Transactions of the Association for Computational Linguistics},
  5:191--204.

\bibitem[{Liu and Forss(2014)}]{Liu2014WebCC}
Shuhua Liu and Thomas Forss. 2014.
\newblock Web content classification based on topic and sentiment analysis of
  text.
\newblock In \emph{KDIR}.

\bibitem[{Liu et~al.(2021)Liu, Zheng, Du, Ding, Qian, Yang, and
  Tang}]{liu2021gpt}
Xiao Liu, Yanan Zheng, Zhengxiao Du, Ming Ding, Yujie Qian, Zhilin Yang, and
  Jie Tang. 2021.
\newblock Gpt understands, too.
\newblock \emph{arXiv:2103.10385}.

\bibitem[{Liu et~al.(2019)Liu, Ott, Goyal, Du, Joshi, Chen, Levy, Lewis,
  Zettlemoyer, and Stoyanov}]{Liu2019RoBERTaAR}
Yinhan Liu, Myle Ott, Naman Goyal, Jingfei Du, Mandar Joshi, Danqi Chen, Omer
  Levy, M.~Lewis, Luke Zettlemoyer, and Veselin Stoyanov. 2019.
\newblock Roberta: A robustly optimized bert pretraining approach.
\newblock \emph{ArXiv}, abs/1907.11692.

\bibitem[{Long et~al.(2018)Long, Ma, Lu, Xiang, and Huang}]{long2018dual}
Yunfei Long, Mingyu Ma, Qin Lu, Rong Xiang, and Chu-Ren Huang. 2018.
\newblock Dual memory network model for biased product review classification.
\newblock In \emph{WASSA@EMNLP}, pages 140--148.

\bibitem[{Loshchilov and Hutter(2019)}]{Loshchilov2019DecoupledWD}
I.~Loshchilov and F.~Hutter. 2019.
\newblock Decoupled weight decay regularization.
\newblock In \emph{ICLR}.

\bibitem[{Ma et~al.(2017)Ma, Li, Zhang, Wang, and Sun}]{ma2017cascading}
Dehong Ma, Sujian Li, Xiaodong Zhang, Houfeng Wang, and Xu~Sun. 2017.
\newblock Cascading multiway attentions for document-level sentiment
  classification.
\newblock In \emph{IJCNLP}, pages 634--643.

\bibitem[{Mahabadi et~al.(2021)Mahabadi, Henderson, and
  Ruder}]{mahabadi2021compacter}
Rabeeh~Karimi Mahabadi, James Henderson, and Sebastian Ruder. 2021.
\newblock Compacter: Efficient low-rank hypercomplex adapter layers.
\newblock \emph{ArXiv}, abs/2106.04647.

\bibitem[{Majumder et~al.(2019)Majumder, Li, Ni, and
  McAuley}]{Majumder2019GeneratingPR}
Bodhisattwa~Prasad Majumder, Shuyang Li, Jianmo Ni, and Julian McAuley. 2019.
\newblock Generating personalized recipes from historical user preferences.
\newblock In \emph{EMNLP/IJCNLP}.

\bibitem[{McAuley et~al.(2012)McAuley, Leskovec, and
  Jurafsky}]{McAuley2012LearningAA}
Julian McAuley, J.~Leskovec, and Dan Jurafsky. 2012.
\newblock Learning attitudes and attributes from multi-aspect reviews.
\newblock In \emph{ICDM}.

\bibitem[{Mikolov et~al.(2013)Mikolov, Chen, Corrado, and
  Dean}]{Mikolov2013EfficientEO}
Tomas Mikolov, Kai Chen, G.~Corrado, and J.~Dean. 2013.
\newblock Efficient estimation of word representations in vector space.
\newblock In \emph{ICLR}.

\bibitem[{Ni et~al.(2019)Ni, Li, and McAuley}]{ni2019justifying}
Jianmo Ni, Jiacheng Li, and Julian McAuley. 2019.
\newblock Justifying recommendations using distantly-labeled reviews and
  fine-grained aspects.
\newblock In \emph{EMNLP-IJCNLP}, pages 188--197.

\bibitem[{Pfeiffer et~al.(2020)Pfeiffer, R{\"u}ckl{\'e}, Poth, Kamath, Vuli'c,
  Ruder, Cho, and Gurevych}]{Pfeiffer2020AdapterHubAF}
Jonas Pfeiffer, Andreas R{\"u}ckl{\'e}, Clifton Poth, Aishwarya Kamath, Ivan
  Vuli'c, Sebastian Ruder, Kyunghyun Cho, and Iryna Gurevych. 2020.
\newblock Adapterhub: A framework for adapting transformers.
\newblock In \emph{EMNLP}.

\bibitem[{Qiu et~al.(2020)Qiu, Sun, Xu, Shao, Dai, and Huang}]{qiu2020pre}
Xipeng Qiu, Tianxiang Sun, Yige Xu, Yunfan Shao, Ning Dai, and Xuanjing Huang.
  2020.
\newblock Pre-trained models for natural language processing: A survey.
\newblock \emph{Science China Technological Sciences}, pages 1--26.

\bibitem[{Raffel et~al.(2020)Raffel, Shazeer, Roberts, Lee, Narang, Matena,
  Zhou, Li, and Liu}]{raffel2020exploring}
Colin Raffel, Noam Shazeer, Adam Roberts, Katherine Lee, Sharan Narang, Michael
  Matena, Yanqi Zhou, Wei Li, and Peter~J Liu. 2020.
\newblock Exploring the limits of transfer learning with a unified text-to-text
  transformer.
\newblock \emph{JMLR}, 21:1--67.

\bibitem[{Ramage et~al.(2009)Ramage, Hall, Nallapati, and
  Manning}]{Ramage2009LabeledLA}
D.~Ramage, David Hall, Ramesh Nallapati, and Christopher~D. Manning. 2009.
\newblock Labeled lda: A supervised topic model for credit attribution in
  multi-labeled corpora.
\newblock In \emph{EMNLP}.

\bibitem[{Rosen-Zvi et~al.(2004)Rosen-Zvi, Griffiths, Steyvers, and
  Smyth}]{RosenZvi2004TheAM}
M.~Rosen-Zvi, T.~Griffiths, M.~Steyvers, and Padhraic Smyth. 2004.
\newblock The author-topic model for authors and documents.
\newblock In \emph{UAI}.

\bibitem[{Sandhaus(2008)}]{sandhaus2008new}
Evan Sandhaus. 2008.
\newblock The new york times annotated corpus.
\newblock \emph{Linguistic Data Consortium, Philadelphia}, 6(12):e26752.

\bibitem[{Schick and Sch{\"u}tze(2021)}]{Schick2021ItsNJ}
Timo Schick and Hinrich Sch{\"u}tze. 2021.
\newblock It’s not just size that matters: Small language models are also
  few-shot learners.
\newblock In \emph{NAACL}.

\bibitem[{Sennrich et~al.(2016)Sennrich, Haddow, and
  Birch}]{Sennrich2016ControllingPI}
Rico Sennrich, B.~Haddow, and Alexandra Birch. 2016.
\newblock Controlling politeness in neural machine translation via side
  constraints.
\newblock In \emph{NAACL}.

\bibitem[{Tang et~al.(2015)Tang, Qin, and Liu}]{tang2015learning}
Duyu Tang, Bing Qin, and Ting Liu. 2015.
\newblock Learning semantic representations of users and products for document
  level sentiment classification.
\newblock In \emph{ACL-IJCNLP}, pages 1014--1023.

\bibitem[{Tay et~al.(2019)Tay, Zhang, Luu, Rao, Zhang, Wang, Fu, and
  Hui}]{Tay2019LightweightAE}
Yi~Tay, A.~Zhang, Anh~Tuan Luu, J.~Rao, Shuai Zhang, Shuohang Wang, Jie Fu, and
  S.~C. Hui. 2019.
\newblock Lightweight and efficient neural natural language processing with
  quaternion networks.
\newblock In \emph{ACL}.

\bibitem[{Vaswani et~al.(2017)Vaswani, Shazeer, Parmar, Uszkoreit, Jones,
  Gomez, Kaiser, and Polosukhin}]{Vaswani2017AttentionIA}
Ashish Vaswani, Noam~M. Shazeer, Niki Parmar, Jakob Uszkoreit, Llion Jones,
  Aidan~N. Gomez, Lukasz Kaiser, and Illia Polosukhin. 2017.
\newblock Attention is all you need.
\newblock In \emph{NIPS}.

\bibitem[{Wan and McAuley(2018)}]{Wan2018ItemRO}
Mengting Wan and Julian McAuley. 2018.
\newblock Item recommendation on monotonic behavior chains.
\newblock In \emph{RecSys}.

\bibitem[{Wang et~al.(2018)Wang, Singh, Michael, Hill, Levy, and
  Bowman}]{Wang2018GLUEAM}
Alex Wang, Amanpreet Singh, Julian Michael, Felix Hill, Omer Levy, and
  Samuel~R. Bowman. 2018.
\newblock Glue: A multi-task benchmark and analysis platform for natural
  language understanding.
\newblock In \emph{ICLR}.

\bibitem[{Wang et~al.(2019)Wang, Wu, Mou, Li, and Chao}]{Wang2019HarnessingPN}
Yunli Wang, Yu~Wu, Lili Mou, Zhoujun Li, and Wen-Han Chao. 2019.
\newblock Harnessing pre-trained neural networks with rules for formality style
  transfer.
\newblock In \emph{EMNLP}.

\bibitem[{Wolf et~al.(2020)Wolf, Debut, Sanh, Chaumond, Delangue, Moi, Cistac,
  Rault, Louf, Funtowicz, and Brew}]{Wolf2020TransformersSN}
Thomas Wolf, Lysandre Debut, Victor Sanh, Julien Chaumond, Clement Delangue,
  Anthony Moi, Pierric Cistac, Tim Rault, R'emi Louf, Morgan Funtowicz, and
  Jamie Brew. 2020.
\newblock Transformers: State-of-the-art natural language processing.
\newblock In \emph{EMNLP}.

\bibitem[{Wu et~al.(2018)Wu, Dai, Yin, Huang, and Chen}]{wu2018improving}
Zhen Wu, Xin-Yu Dai, Cunyan Yin, Shujian Huang, and Jiajun Chen. 2018.
\newblock Improving review representations with user attention and product
  attention for sentiment classification.
\newblock In \emph{AAAI}.

\bibitem[{Xu et~al.(2019)Xu, Hu, Leskovec, and Jegelka}]{Xu2019HowPA}
Keyulu Xu, Weihua Hu, J.~Leskovec, and S.~Jegelka. 2019.
\newblock How powerful are graph neural networks?
\newblock In \emph{ICLR}.

\bibitem[{Yang et~al.(2017)Yang, Mei, Ji, Zhao, Zhao, and
  Chen}]{Yang2017IdentifyingAT}
Min Yang, Jincheng Mei, Heng Ji, Wei Zhao, Zhou Zhao, and Xiaojun Chen. 2017.
\newblock Identifying and tracking sentiments and topics from social media
  texts during natural disasters.
\newblock In \emph{EMNLP}.

\bibitem[{Yang et~al.(2018)Yang, Sun, Li, and Ma}]{Yang2018AutomaticAP}
Pengcheng Yang, Xu~Sun, Wei Li, and Shuming Ma. 2018.
\newblock Automatic academic paper rating based on modularized hierarchical
  convolutional neural network.
\newblock In \emph{ACL}.

\bibitem[{Zhang et~al.(2021)Zhang, Tay, Zhang, Chan, Luu, Hui, and
  Fu}]{Zhang2021BeyondFL}
A.~Zhang, Yi~Tay, Shuai Zhang, Alvin Chan, A.~Luu, S.~C. Hui, and Jie Fu. 2021.
\newblock Beyond fully-connected layers with quaternions: Parameterization of
  hypercomplex multiplications with 1/n parameters.
\newblock In \emph{ICLR}.

\bibitem[{Zhao and Mao(2017)}]{Zhao2017TopicAwareDC}
Rui Zhao and K.~Mao. 2017.
\newblock Topic-aware deep compositional models for sentence classification.
\newblock \emph{IEEE/ACM Transactions on Audio, Speech, and Language
  Processing}, 25:248--260.

\bibitem[{Zhao et~al.(2021)Zhao, Wallace, Feng, Klein, and
  Singh}]{Zhao2021CalibrateBU}
Tony Zhao, Eric Wallace, Shi Feng, D.~Klein, and Sameer Singh. 2021.
\newblock Calibrate before use: Improving few-shot performance of language
  models.
\newblock In \emph{ICML}.

\bibitem[{Zhu et~al.(2015)Zhu, Kiros, Zemel, Salakhutdinov, Urtasun, Torralba,
  and Fidler}]{zhu2015aligning}
Yukun Zhu, Ryan Kiros, Rich Zemel, Ruslan Salakhutdinov, Raquel Urtasun,
  Antonio Torralba, and Sanja Fidler. 2015.
\newblock Aligning books and movies: Towards story-like visual explanations by
  watching movies and reading books.
\newblock In \emph{ICCV}, pages 19--27.

\end{thebibliography}
\bibliographystyle{acl_natbib}

\clearpage

\appendix
\section{Appendix}

\subsection{Training Configurations for Reproducibility}

Our model is implemented in Python 3, and mainly uses the
following dependencies: \texttt{torch}
as the machine learning library, \texttt{nltk}
for text preprocessing, \texttt{transformers}
for
their BERT implementation, and \texttt{numpy}
for high-level mathematical operations in CPU. During our experiments, we used machines with a
single GeForce GTX 1080Ti GPU, 4 CPUs and 16GB of
RAMs. The training times for all datasets are less than a day. 
The total number of parameters depends on the number of attributes, each of which has their own attribute-specific adapters.
In our experiments, excluding the embedding matrices and classifiers that vary a lot across datasets and tasks,
BERT-base with \textsc{Injectors} can have a total of 105M parameters with 19M (18\%) trained for tasks with two attributes, or a total of 121M parameters with 36M (29\%) trained for tasks with four attributes.
Using the accuracy of the model on the development set, we tuned the learning rate (from $1e-6$, $3e-5$, $1e-5$, and $3e-4$), the adapter size (from $32$, $48$, $64$, and $128$), and the hypercomplex dimensions (from 2, 4, 6, and 8).

\subsection{Descriptions of Newly Introduced Datasets}

This section describes how we procured the three datasets we introduce in this paper:

\begin{enumerate}
    \item \textbf{Food.com}: We used the dataset gathered in \citet{Majumder2019GeneratingPR}, which was used as a personalized recipe generation dataset. We repurposed the dataset for a new classification task and used the recipes as input text and the duration (in minutes) as output class. We removed instances with outliers: (1) recipes that took less than 5 minutes and more than 150 minutes; (2) recipes with more than 500 tokens or less than 10 tokens; and (3) tags with more than 50 labels. We also removed from the attribute vocabulary tags that explicitly indicate the recipe duration (e.g., \texttt{60-minutes-or-less}) and those that are used on almost all instances (e.g., \texttt{time-to-make}). We shuffled the data and used 10\% each for the development and test sets, and the remaining 80\% for the training set.
    \item \textbf{Goodreads}: We used the review corpus gathered in \citet{Wan2018ItemRO}, which was also used for spoiler detection. Since the split is unfortunately not publicly shared, we created our own split. We first removed very short (less than 32 tokens) and very long (more than 256) reviews as they were outliers. We then divided the data into three splits, with two 10K splits as the development and test sets, and the remaining split as the training set.
    \item \textbf{Beeradvocate}: We used the review corpus gathered in \citet{McAuley2012LearningAA}. We removed outliers and split the dataset into three using the same method we did with Goodreads.
\end{enumerate}

\subsection{Parameter Analysis of Weight-based Injection}

Recall that we define $\mathbf{W}_{z_j} \in \mathbb{R}^{d_h \times d_a}$ as follows:
\begin{equation}
    \mathbf{W}_{z_j} = g_{weight}(\mathbf{z}_j) + \mathbf{C}_{weight}
\end{equation}

In a naive setting, we can trivially use a projection function as our $g_{weight}$, which would linearly transform $\mathbf{z}_j \in \mathbb{R}^{d_z}$ into the shape $d_h \times d_a$. This would need a weight tensor of size $d_z \times d_h \times d_a$, which can be prohibitively large.
This parameter dominates all the other parameters in the module, thus the overall parameter of the naive method is $\mathcal{O}(d_z * d_h * d_a)$.

Our parameter-saving methods remove this large tensor, but instead use three smaller parameters in hypercomplex space: the transform function $\sigma_o(\cdot)$ that is basically a linear transformation with a projection matrix of size $d_z \times d_a$ (Eq. \ref{eq:hypercomplex}), the vector $\mathbf{s}_o$ of size $d_h / O^2$, and the matrix $\mathbf{A}_o$ of size $O \times O$. Since we have $O$ dimensions in our hypercomplex space, we have a total of $O * (d_z * d_a + d_h / O^2 + O^2)$, which we can reduce as follows:
\begin{align}
    &O * (d_z * d_a + d_h / O^2 + O^2)\nonumber \\
    &=  O * d_z * d_a + d_h / O + O^3\nonumber \\
    &\approx  O * d_z * d_a\nonumber \\
    &\approx  d_z * d_a
\end{align}
given that $O^3 \ll O*d_z*d_a$ and that we can treat $O$ as a constant ($O=4$ in our experiment). Thus the overall parameter when using our parameter-saving method is $\mathcal{O}(d_z * d_a)$. We emphasize that this is a huge improvement since the PLM hidden size $d_h$ is usually the largest dimension.

The output weight $\mathbf{W}_{z_j}$ has a rank $r$ of at most $O^2+1$, i.e., (1) the low-rank method (Eq. \ref{eq:low-rank}) outputs a matrix of rank $r=1$; (2) the Kronecker product (Eq. \ref{eq:phm1}) returns a matrix of rank $r=O$; and finally, (3) the sum of multiple matrices (Eq. \ref{eq:phm2}) has a rank $r \leq O^2$.

\subsection{Full Ablation Studies}

Table \ref{tab:full-ablation} reports various ablation studies on all datasets which assess the contribution of the different components of our models. We can see very similar observations in this table and the table shown in the main text of our paper. 

\begin{table*}[t]
    \small
	\centering
	\begin{tabular}{@{}lcccccccc@{}}
		\thickhline
		\multicolumn{1}{c}{Model} &  Yelp 2013    & Yelp 2014    & IMDB & AAPR & PolMed & Food.com & Goodreads & BeerAdvocate \\
		\thickhline
		$\mathcal{B}$ + \textsc{Injectors} & {70.76} & {71.35} & 55.13 & {67.10} & {47.27} & {45.01} & {57.78} & {57.69} \\ 
		\hline
		\quad -- bias injection & 70.33 & 71.24 & 55.06 & 66.45 & 46.67 & 44.86 & 57.06 & 57.67 \\
		\quad -- weight injection & 70.51 & 71.27 & 54.82 & 66.85 & 45.66 & 44.74 & 57.61 & 57.29 \\
		\quad -- task adapter & 69.21 & 69.68 & 54.03 & 65.55 & 46.89 & 44.30 & 56.57 & 57.28 \\ \hline
		\quad -- attribute drop & 69.29 & 70.94 & 54.01 & 65.55 & 46.33 & 44.48 & 56.62 & 57.41 \\
		\quad -- post-aggregation & 70.76 & 71.35 & 55.13 & 64.42 & 47.27 & 43.30 & 57.78 & 57.69  \\ \hline
		\quad -- low-rank & OOM & OOM & OOM & OOM & OOM & OOM & OOM & OOM \\
		\quad -- PHM & 68.27 & 69.05 & 54.15 & 65.15 & 46.16 & 43.89 & 55.99 & 56.51 \\
		\thickhline
	\end{tabular}%
	\caption{Performance of \textsc{Injectors} and versions thereof without some of our proposed components (second block), training mechanisms (third block), and parameter-saving methods (fourth block). OOM denotes the model does not run on our experimental setup due to out of memory error.}
	\label{tab:full-ablation}%
\end{table*}%

\subsection{Performance Plots per Sparsity Level}

\begin{figure*}[t]
    \centering
    \begin{tabular}{@{}c@{}c@{}c@{}} \hline
     \multicolumn{3}{@{}c@{}}{\textbf{Food.com}} \\
    \includegraphics[width=0.33\textwidth]{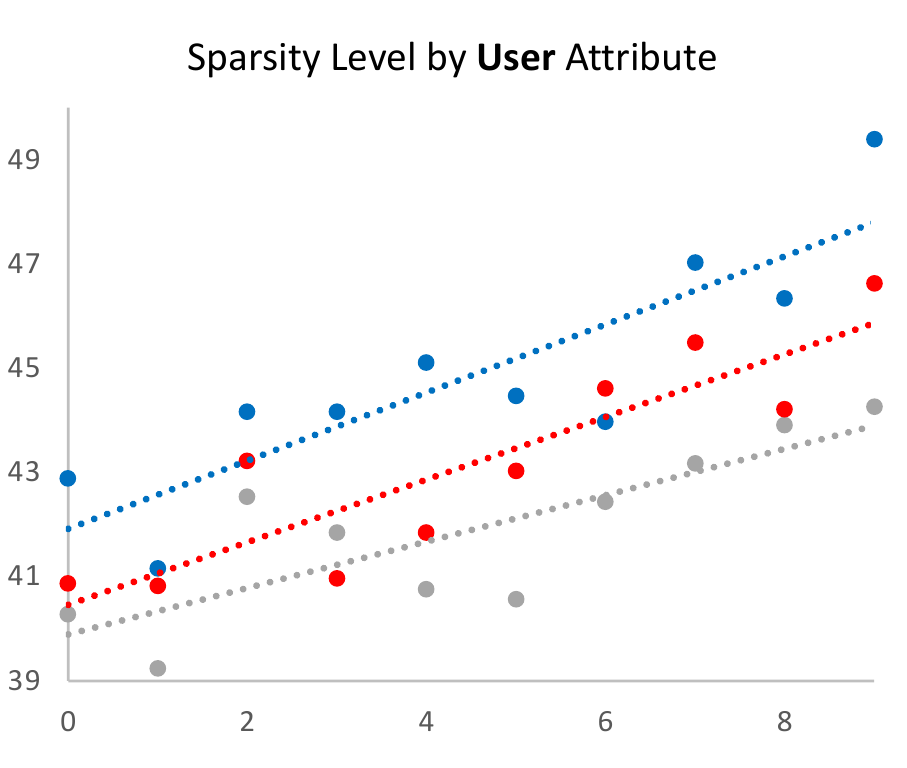} &
    \includegraphics[width=0.33\textwidth]{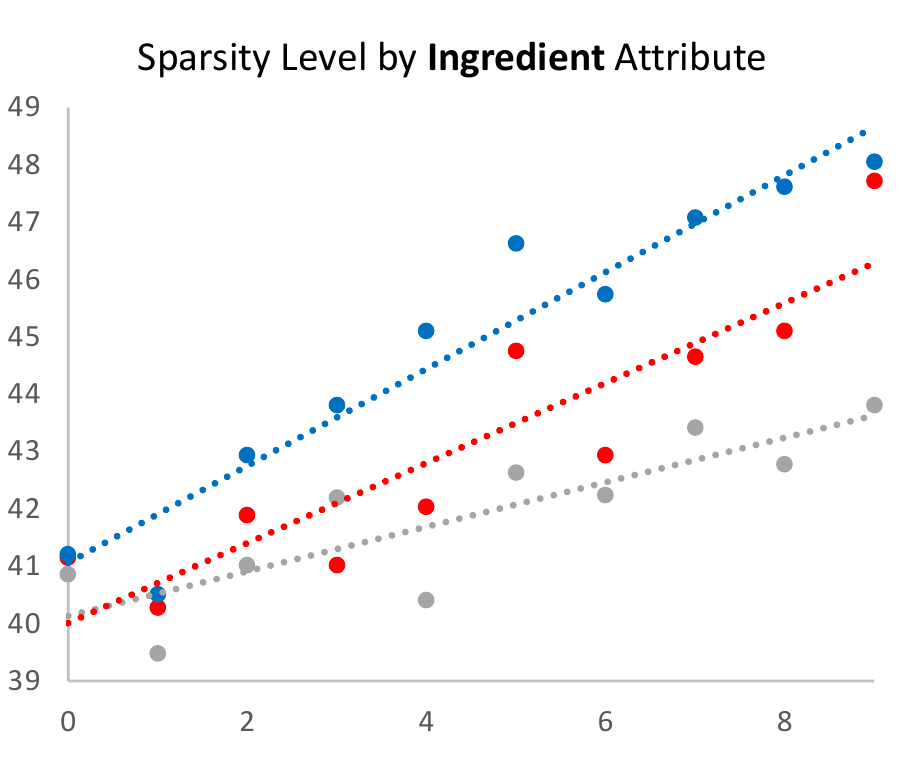} &
    \includegraphics[width=0.33\textwidth]{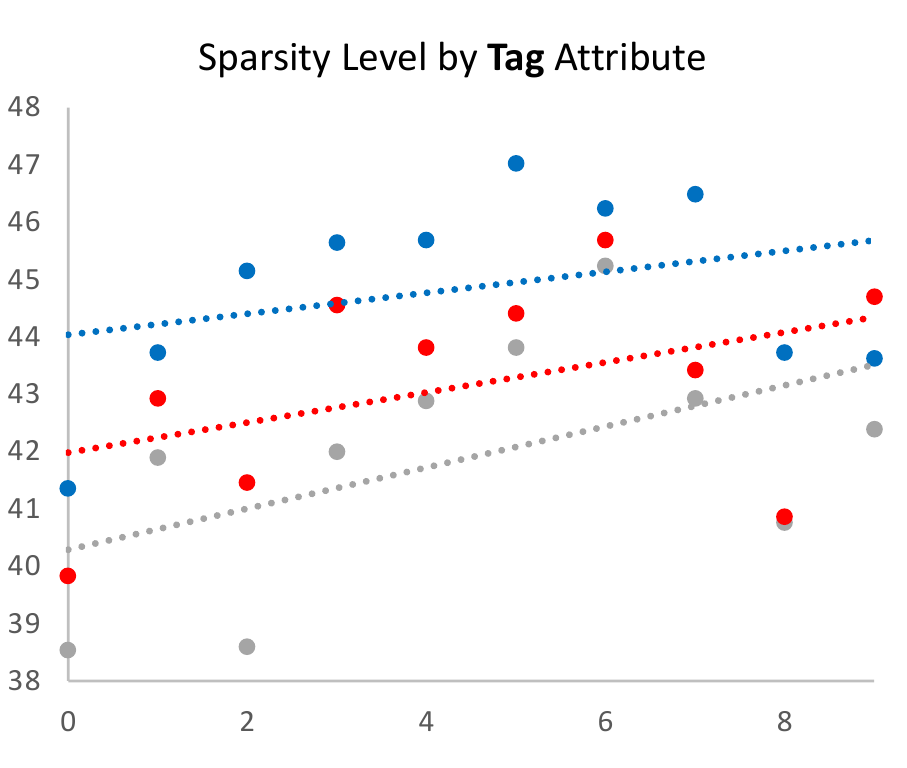} \\
    \hline
    \multicolumn{3}{@{}c@{}}{\textbf{Goodreads}} \\
    \includegraphics[width=0.33\textwidth]{good1.pdf} &
    \includegraphics[width=0.33\textwidth]{good2.pdf} &
    \includegraphics[width=0.33\textwidth]{good3.pdf} \\
     \hline
     \multicolumn{3}{@{}c@{}}{\textbf{BeerAdvocate}} \\
    \includegraphics[width=0.33\textwidth]{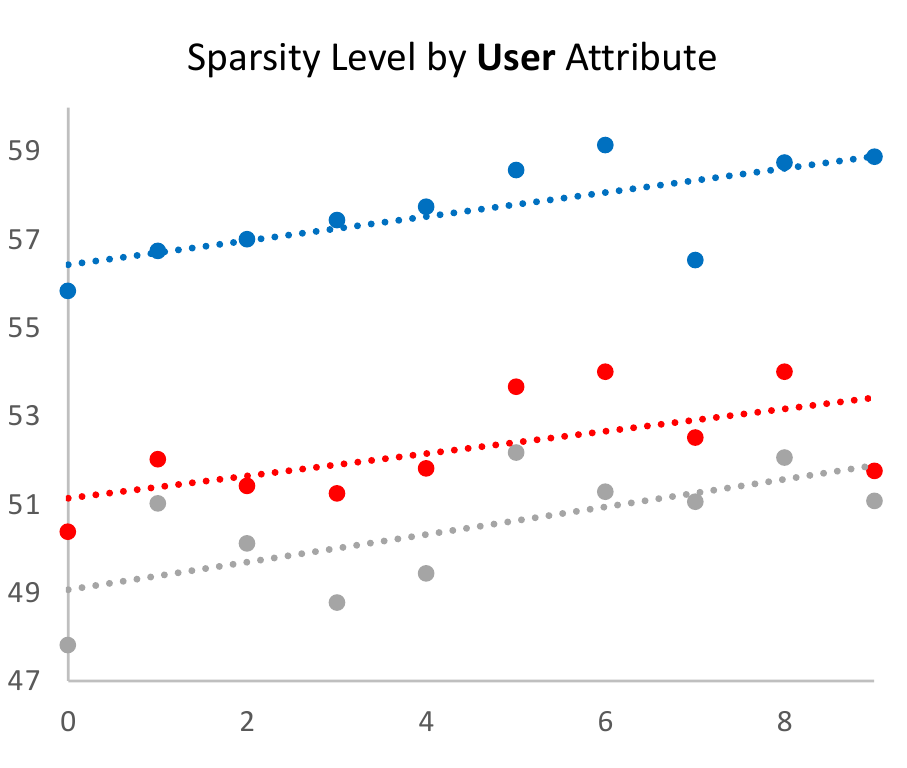} &
    \includegraphics[width=0.33\textwidth]{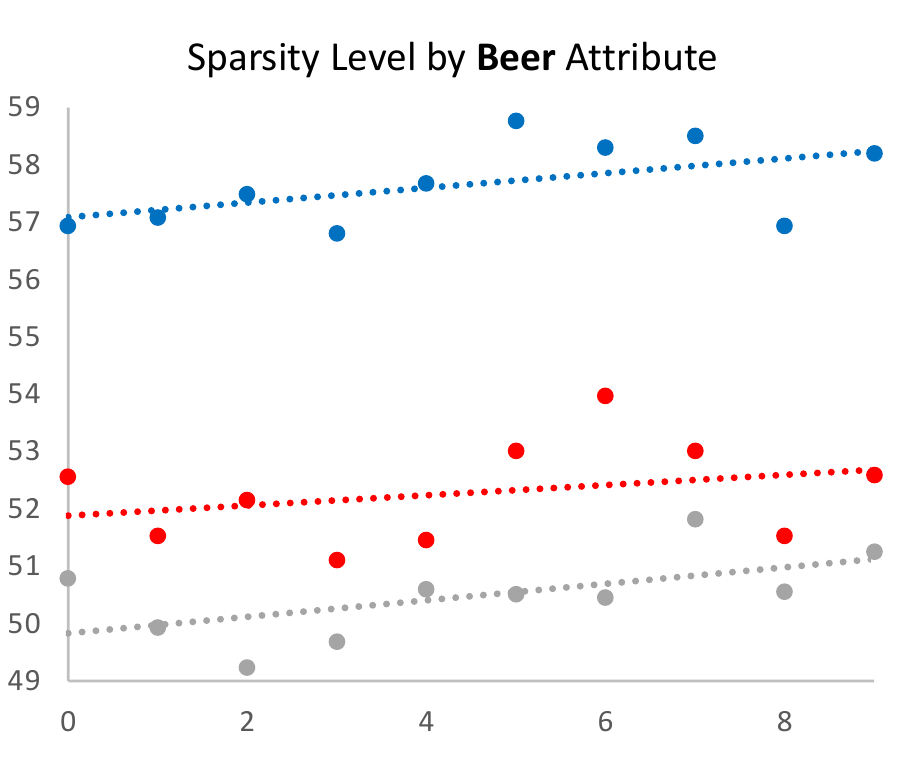} &
    \includegraphics[width=0.33\textwidth]{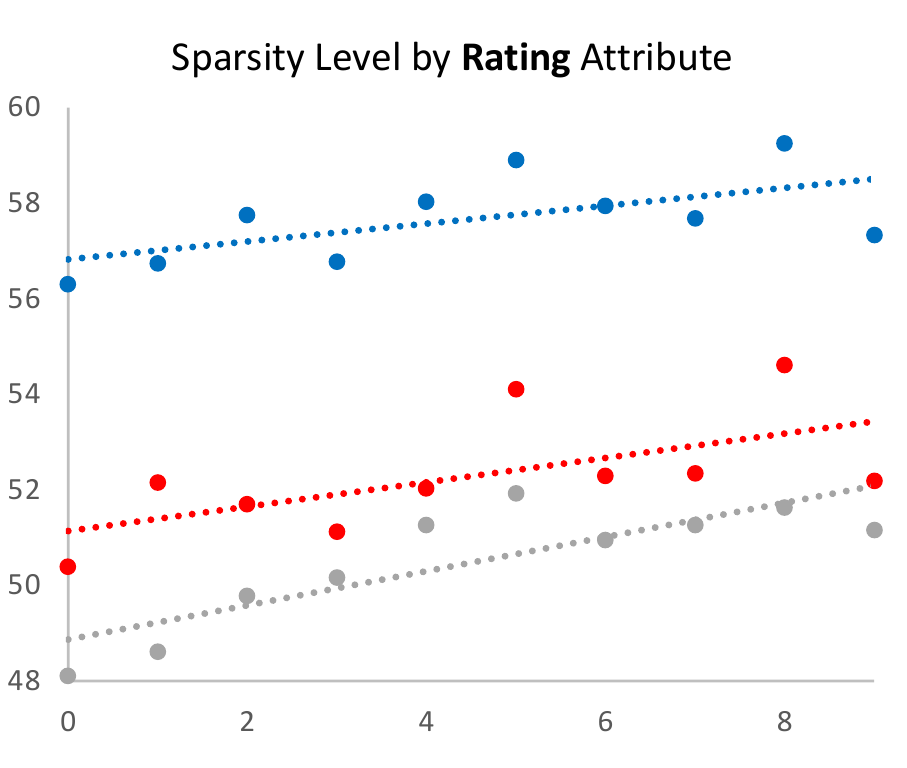} \\
     \hline
    \end{tabular}
    \caption{Performance plots per sparsity level of BERT-base (gray), CHIM (red), and \textsc{Injectors} (blue) for each attribute in the Food.com, Goodreads, and Beeradvocate datasets. The x-axis with a lower value has a higher sparsity level (0 is the most sparse).}
    \label{fig:sparse-graph1}
\end{figure*}

\begin{figure*}[t]
    \centering
    \begin{tabular}{@{}c@{}c@{}c@{}c@{}} \hline
    \multicolumn{2}{@{}c@{}}{\textbf{Yelp2013}} &
    \multicolumn{2}{@{}c@{}}{\textbf{Yelp2014}}\\
    \includegraphics[width=0.25\textwidth]{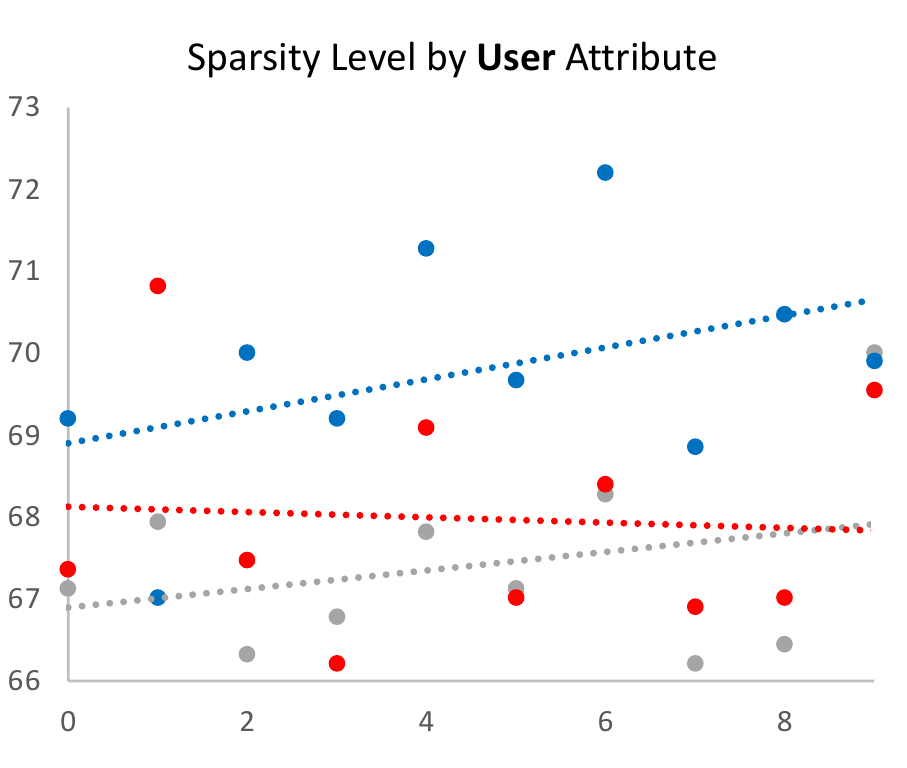} &
    \includegraphics[width=0.25\textwidth]{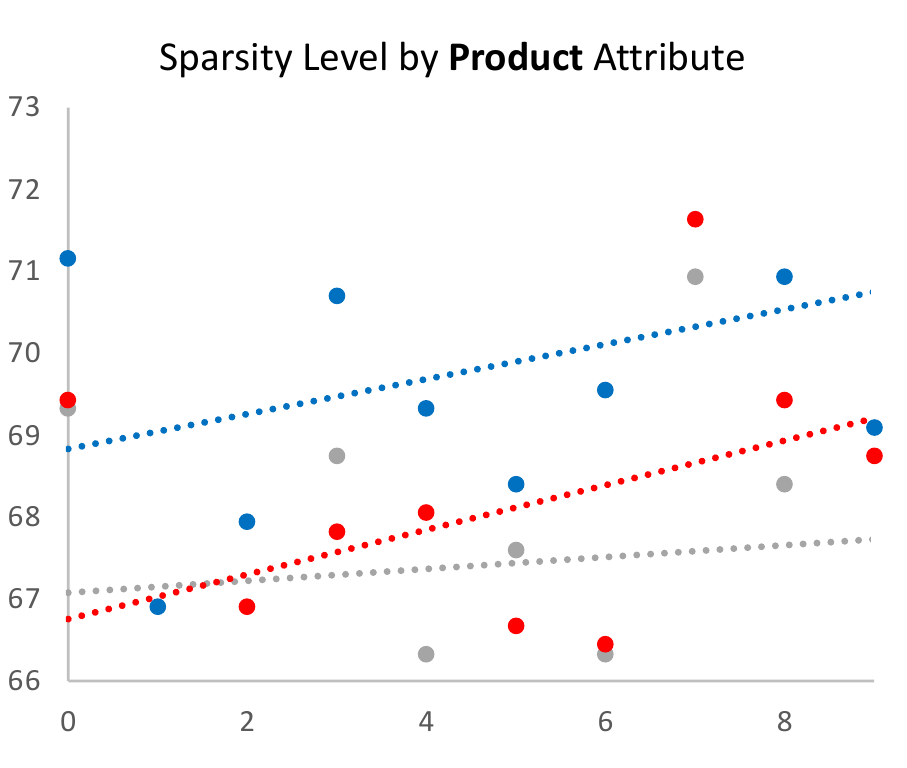} &
    \includegraphics[width=0.25\textwidth]{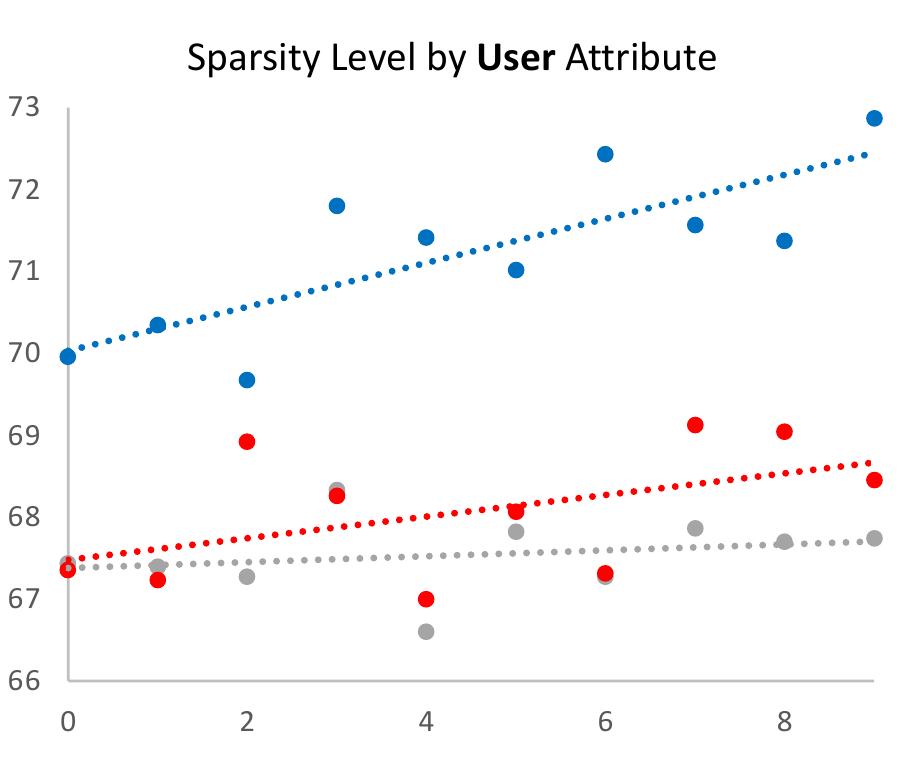} &
    \includegraphics[width=0.25\textwidth]{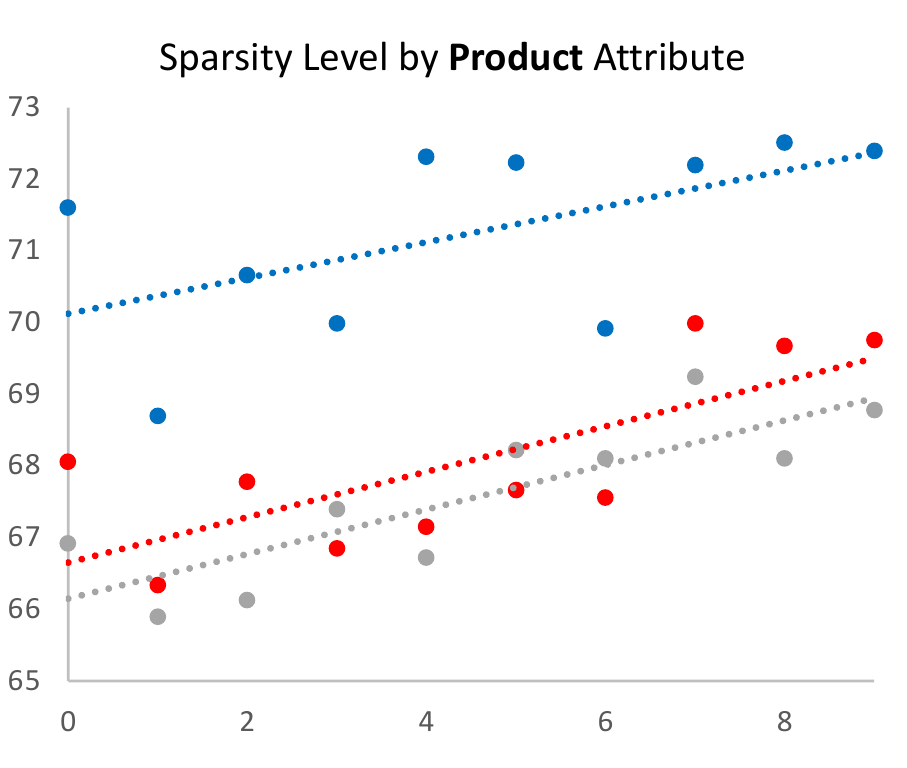} \\
     \hline
     \multicolumn{2}{@{}c@{}}{\textbf{IMDB}} &
    \multicolumn{2}{@{}c@{}}{\textbf{AAPR}}\\
    \includegraphics[width=0.25\textwidth]{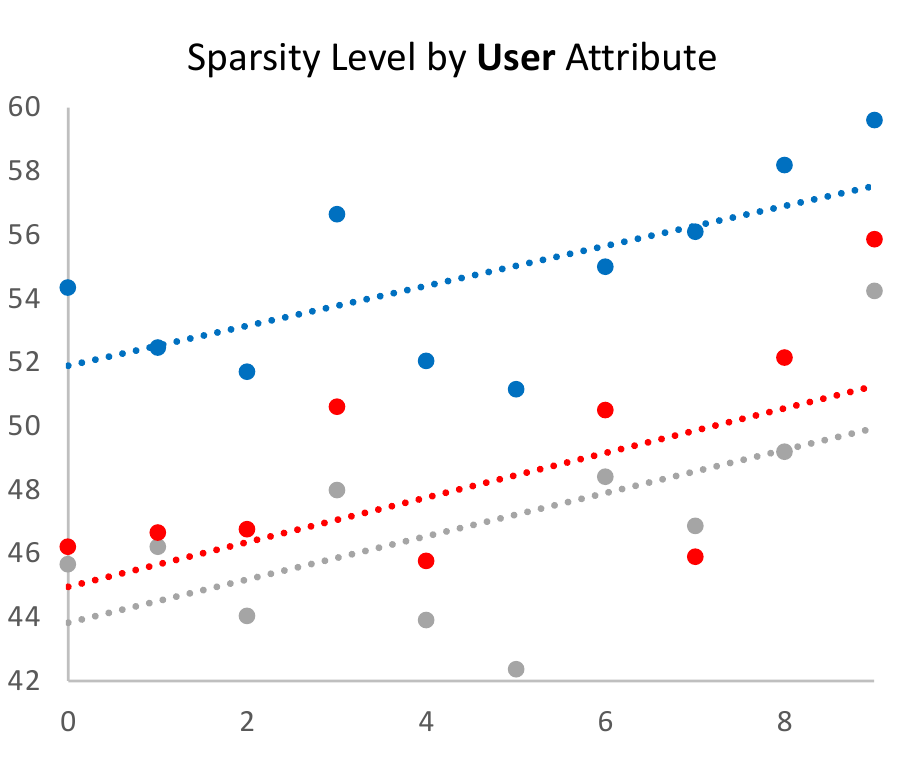} &
    \includegraphics[width=0.25\textwidth]{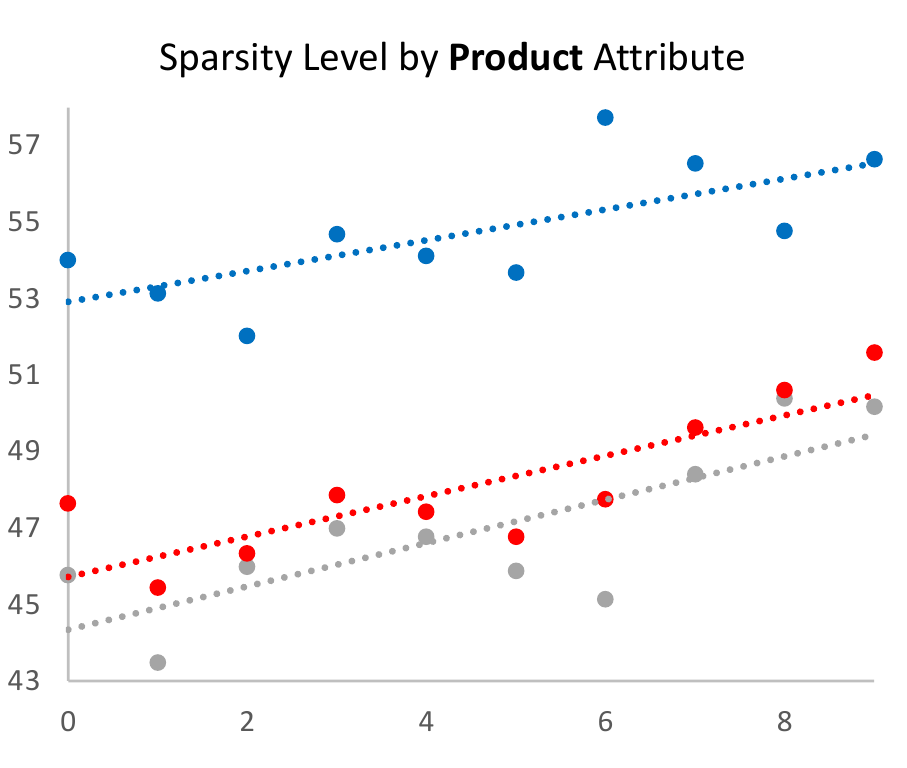} &
    \includegraphics[width=0.25\textwidth]{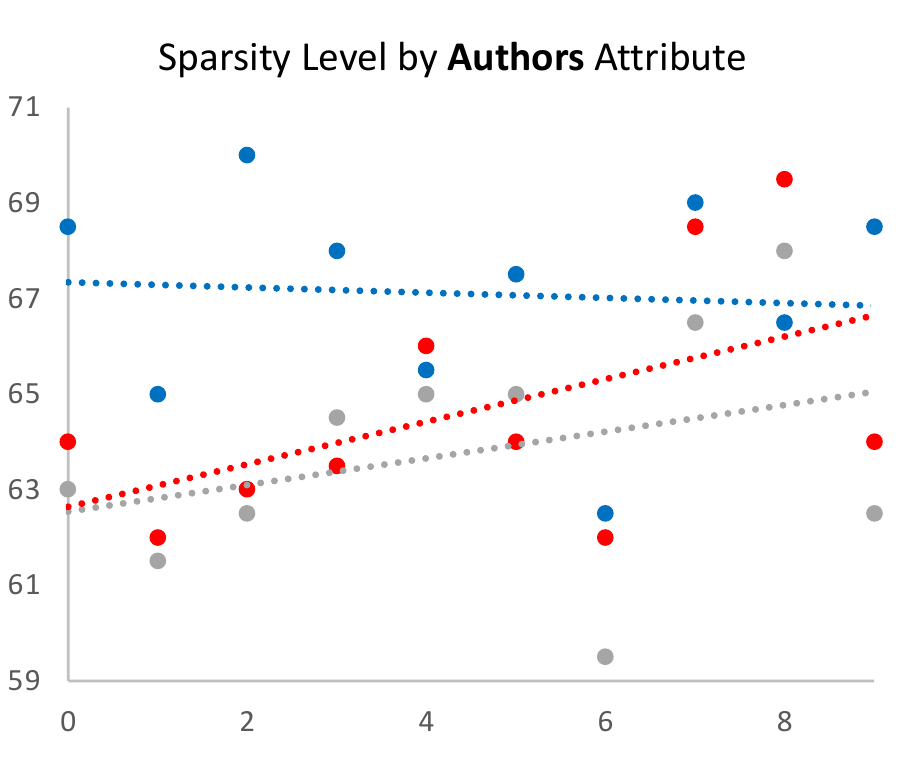} &
    \includegraphics[width=0.25\textwidth]{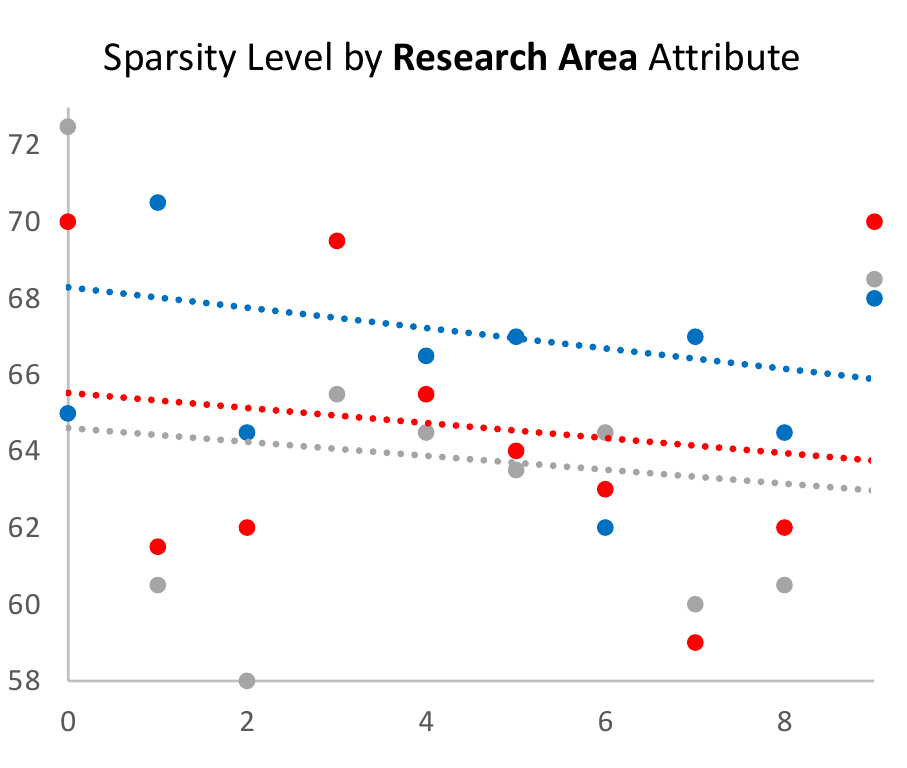} \\
     \hline
     \multicolumn{4}{@{}c@{}}{\textbf{PolMed}} \\
    \includegraphics[width=0.25\textwidth]{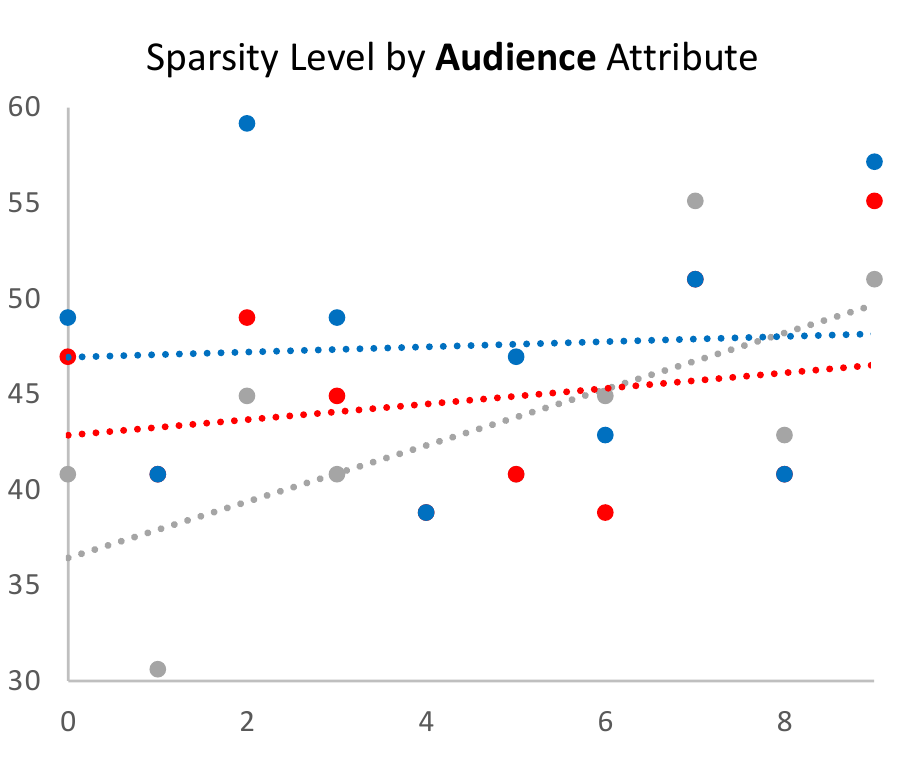} &
    \includegraphics[width=0.25\textwidth]{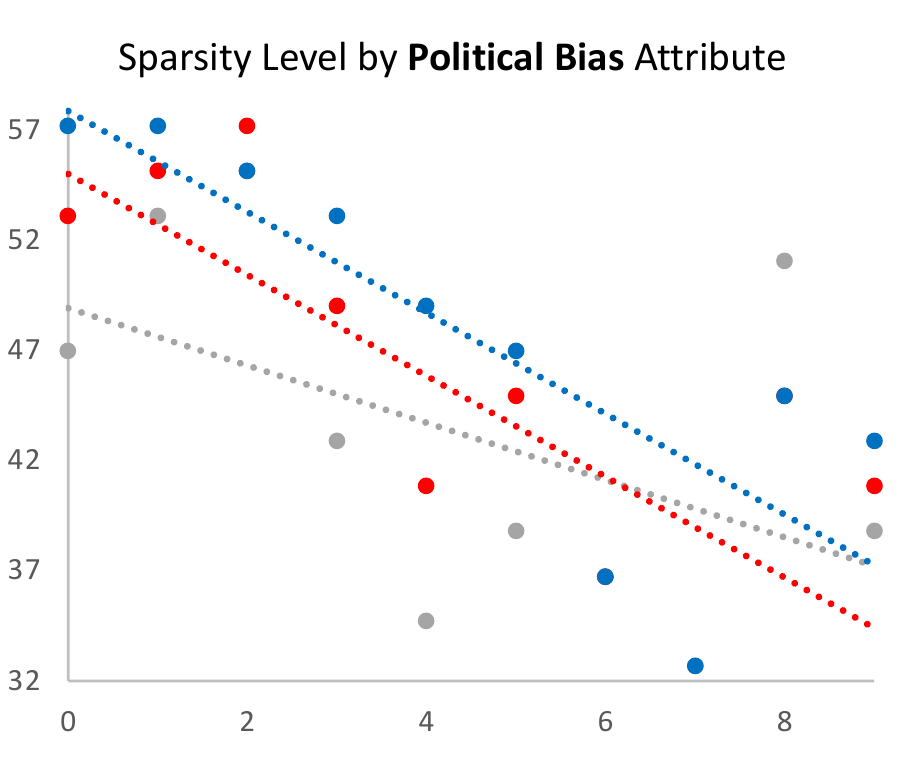} &
    \includegraphics[width=0.25\textwidth]{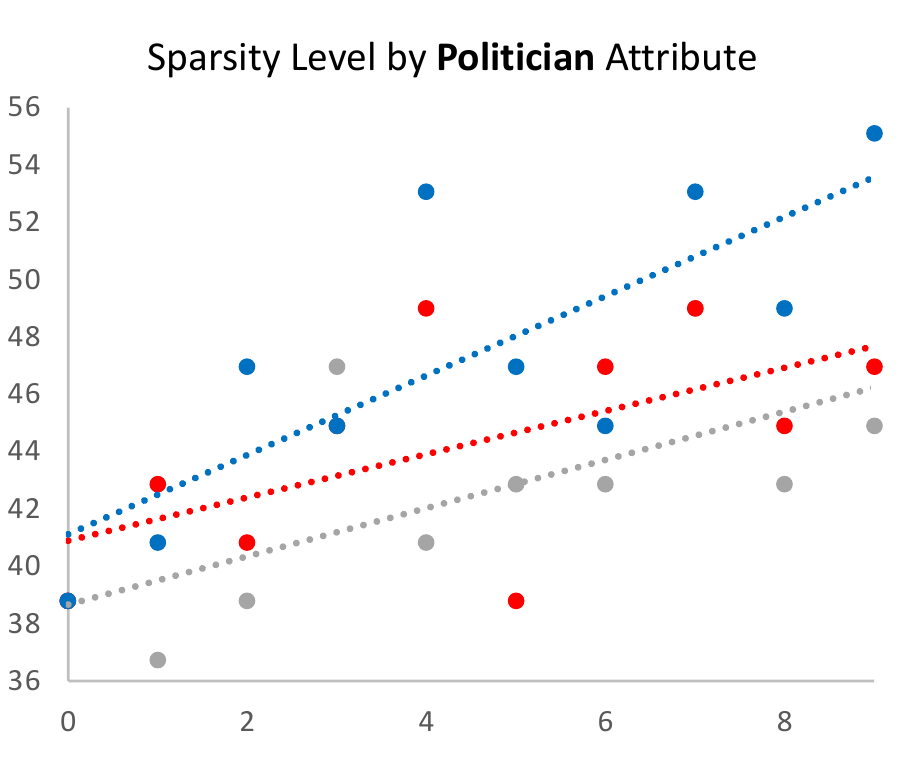} &
    \includegraphics[width=0.25\textwidth]{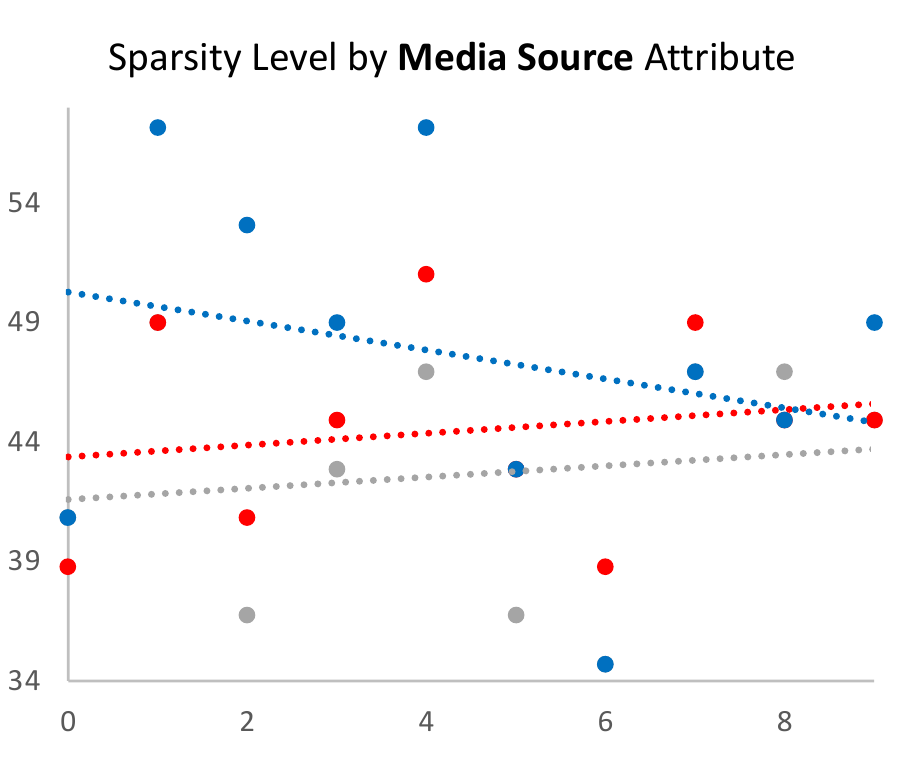} \\
     \hline
    \end{tabular}
    \caption{Performance plots per sparsity level of BERT-base (gray), CHIM (red), and \textsc{Injectors} (blue) for each attribute in the Yelp2013, Yelp2014, IMDB, AAPR, and PolMed datasets. The x-axis with a lower value has a higher sparsity level (0 is the most sparse).}
    \label{fig:sparse-graph2}
\end{figure*}

Finally, we show the performance plots per sparsity level for all datasets in Figures \ref{fig:sparse-graph1}--\ref{fig:sparse-graph2}. Overall, when fitting the plots into a line, \textsc{Injectors} outperform CHIM on all datasets and sparsity levels. For very sparse attributes (e.g., AAPR authors, Goodreads user, etc.), we can clearly see that the increase in performance is substantially larger in the sparser levels.

\end{document}